\documentclass{article} 
\usepackage[preprint]{neurips_2024}

\input{mymath.tex}

\usepackage{hyperref}
\usepackage{url}
\usepackage{amsmath}
\usepackage{amssymb}
\usepackage{bm,bbm}
\usepackage[makeroom]{cancel}
\usepackage{adjustbox}
\usepackage{xspace}
\usepackage{color}
\usepackage{xcolor}
\usepackage{colortbl}
\usepackage{subcaption}
\usepackage{multirow}
\usepackage{adjustbox}
\usepackage{booktabs}
\usepackage{makecell}
\usepackage{wrapfig}

\usepackage{utfsym} 
\newcommand{\heart}{$\;\!$\usym{2665}}

\newcommand{\ourMethod}[0]{SVE-Math} 
\newcommand{\ourgip}[0]{GeoGLIP} 
\definecolor{citecolor}{HTML}{2980b9}
\definecolor{linkcolor}{HTML}{c0392b}
\definecolor{backred}{RGB}{255, 190, 190}
\definecolor{backblue}{RGB}{210, 230, 250}
\definecolor{pink}{RGB}{240, 81, 121}
\definecolor{green}{RGB}{69, 189, 155}
\definecolor{yellow}{RGB}{253, 207, 110}
\definecolor{lightred}{RGB}{254, 129, 125}
\definecolor{lightblue}{RGB}{129, 184, 223}
\newcolumntype{C}[1]{>{\centering\arraybackslash}m{#1}}
\makeatletter
  \newcommand\figcaption{\def\@captype{figure}\caption}
  \newcommand\tabcaption{\def\@captype{table}\caption}
\makeatother
\title{Open Eyes, Then Reason: Fine-grained Visual Mathematical Understanding in MLLMs}

\author{%
Shan Zhang$^{1}$ \quad Aotian Chen$^{2}$ \quad Yanpeng Sun$^{3}$ \quad Jindong Gu$^{4}$ \quad  Yi-Yu Zheng$^{5}$\\ $\!\!\!$\textbf{Piotr Koniusz}$^{6}$\quad \textbf{Kai Zou}$^{5}$ \quad \textbf{Anton van den Hengel}$^{1 *}$ \quad \textbf{Yuan Xue}$^{7}$\thanks{The corresponding author.}\\
$^{1}$Australian Institute for Machine Learning, University of Adelaide \\
$^2$Georgia Institute of Technology  \quad $^3$Nanjing University of Science and Technology\\
$^{4}$University of Oxford \quad $^{5}$NetMind.ai \quad $^6$Data61$\!${\color{red}$\heart$}CSIRO \quad  $^{7}$The Ohio State University\\
{\tt\small $^{1}$shan.zhang@adelaide.edu.au, $^7$yuan.xue@osumc.edu}
}
\begin{document}

\maketitle

\begin{abstract}

Current multimodal large language models (MLLMs) often underperform on mathematical problem-solving tasks that require fine-grained visual understanding. The limitation is largely attributable to inadequate perception of geometric primitives during image-level contrastive pre-training (\eg, CLIP). While recent efforts to improve math MLLMs have focused on scaling up mathematical visual instruction datasets and employing stronger LLM backbones, they often overlook persistent errors in visual recognition. In this paper, we systematically evaluate the visual grounding capabilities of state-of-the-art MLLMs and reveal a significant negative correlation between visual grounding accuracy and problem-solving performance, underscoring the critical role of fine-grained visual understanding. Notably, advanced models like GPT-4o exhibit a 70\% error rate when identifying geometric entities, highlighting that this remains a key bottleneck in visual mathematical reasoning. To address this, we propose a novel approach, \textit{SVE-Math} (Selective Vision-Enhanced Mathematical MLLM), featuring a geometric-grounded
vision encoder and a feature router that dynamically adjusts the contribution of hierarchical visual feature maps. Our model recognizes accurate visual primitives and generates precise visual prompts tailored to the language model's reasoning needs. In experiments, \ourMethod-Qwen2.5-7B outperforms other 7B models by 15\% on MathVerse and is compatible with GPT-4V on MathVista. Despite being trained on smaller datasets, \ourMethod-7B achieves competitive performance on GeoQA, rivaling models trained on significantly larger datasets. Our findings emphasize the importance of incorporating fine-grained visual understanding into MLLMs and provide a promising direction for future research. Code is available at \href{https://github.com/AI4Math-ShanZhang/SVE-Math}{\color{red}{github.com/AI4Math-ShanZhang/SVE-Math}}.




\end{abstract}

\section{Introduction}
\label{sec:intro}

Visual information plays a crucial role in mathematical problem-solving, where diagrams and visual representations often encapsulate relationships and properties essential for understanding and reasoning. While Large Language Models (LLMs) have demonstrated impressive capabilities in textual mathematical reasoning~\citep{yu2023metamath,ying2024internlm,azerbayev2023llemma}, their proficiency often diminishes when tasks require integrating visual data. The challenge intensifies when precise comprehension of geometric primitives—basic elements such as lines, circles, angles, boundaries, and junctions—is necessary to solve complex mathematical problems.
Recent advancements in Multimodal Large Language Models (MLLMs)~\citep{Chen2022UniGeoUG,liang_unimath,kazemi2023geomverse,gao2023g,zhang2024mavis,shi2024math} have shown promise in addressing visual mathematical reasoning by incorporating both textual and visual inputs. These models typically rely on large-scale mathematical visual instruction datasets~\citep{zhang2024mavis,shi2024math,kazemi2023geomverse}, which require MLLMs~\citep{OpenAI2023ChatGPT,openai2023gpt4v,su2023pandagpt} to generate diverse descriptions for question-answer pairs involving geometric elements.
While these approaches enhance the reasoning capabilities of MLLMs in the mathematical domain, they come with certain limitations. Constructing such datasets is time-consuming, labor-intensive, and requires substantial financial and human resources, often involving the use of advanced models like GPT-4o~\citep{openai2023gpt4v} to generate diverse prompts for synthetic datasets.

\begin{figure}[t]
    \centering
    
    \begin{subfigure}[b]{0.32\linewidth}
        \centering
        \includegraphics[width=\linewidth]{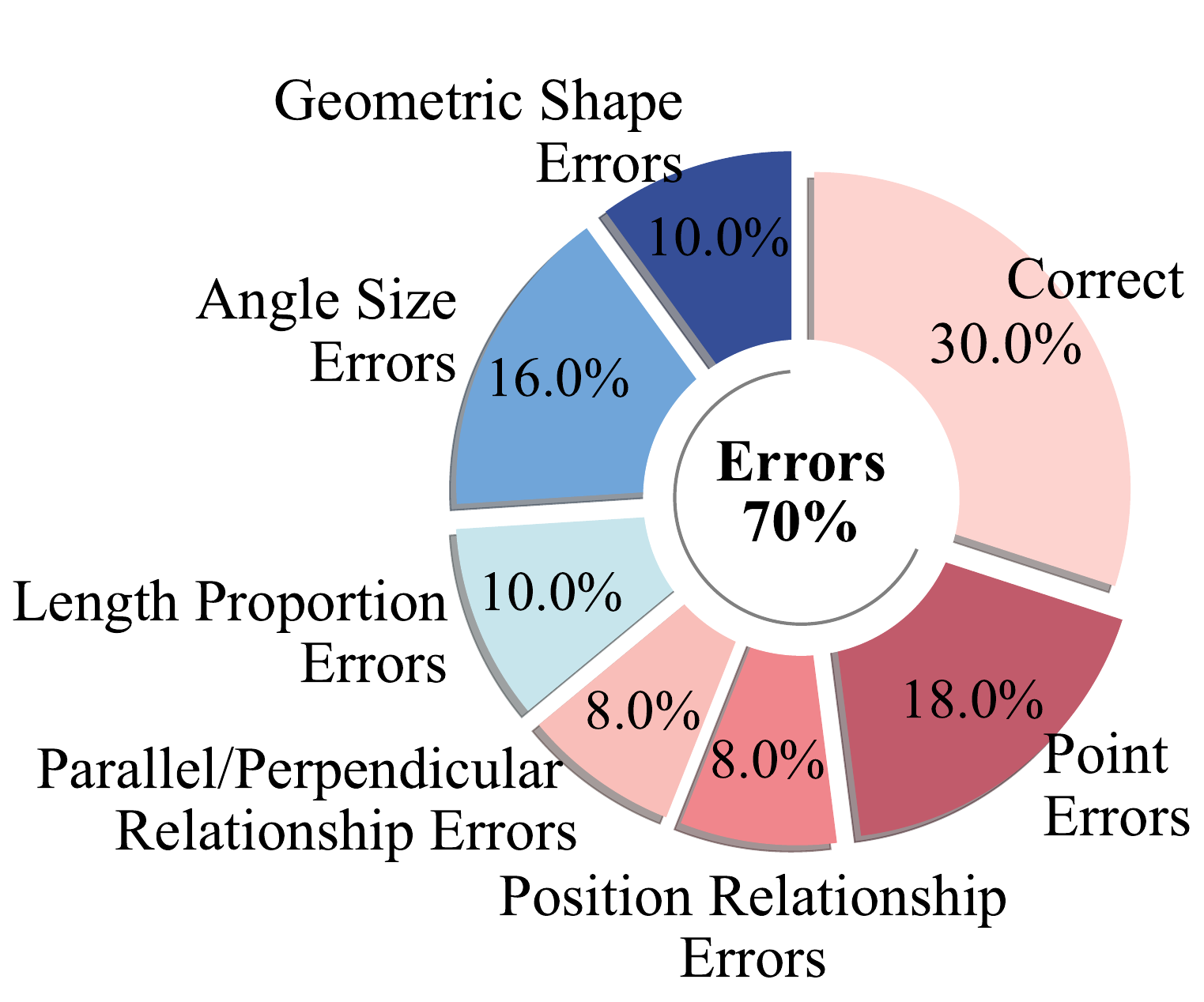}
        \caption{}
        \label{fig:intro_subfig1}
    \end{subfigure}
    \hfill
    \begin{subfigure}[b]{0.32\linewidth}
        \centering
        \includegraphics[width=\linewidth]{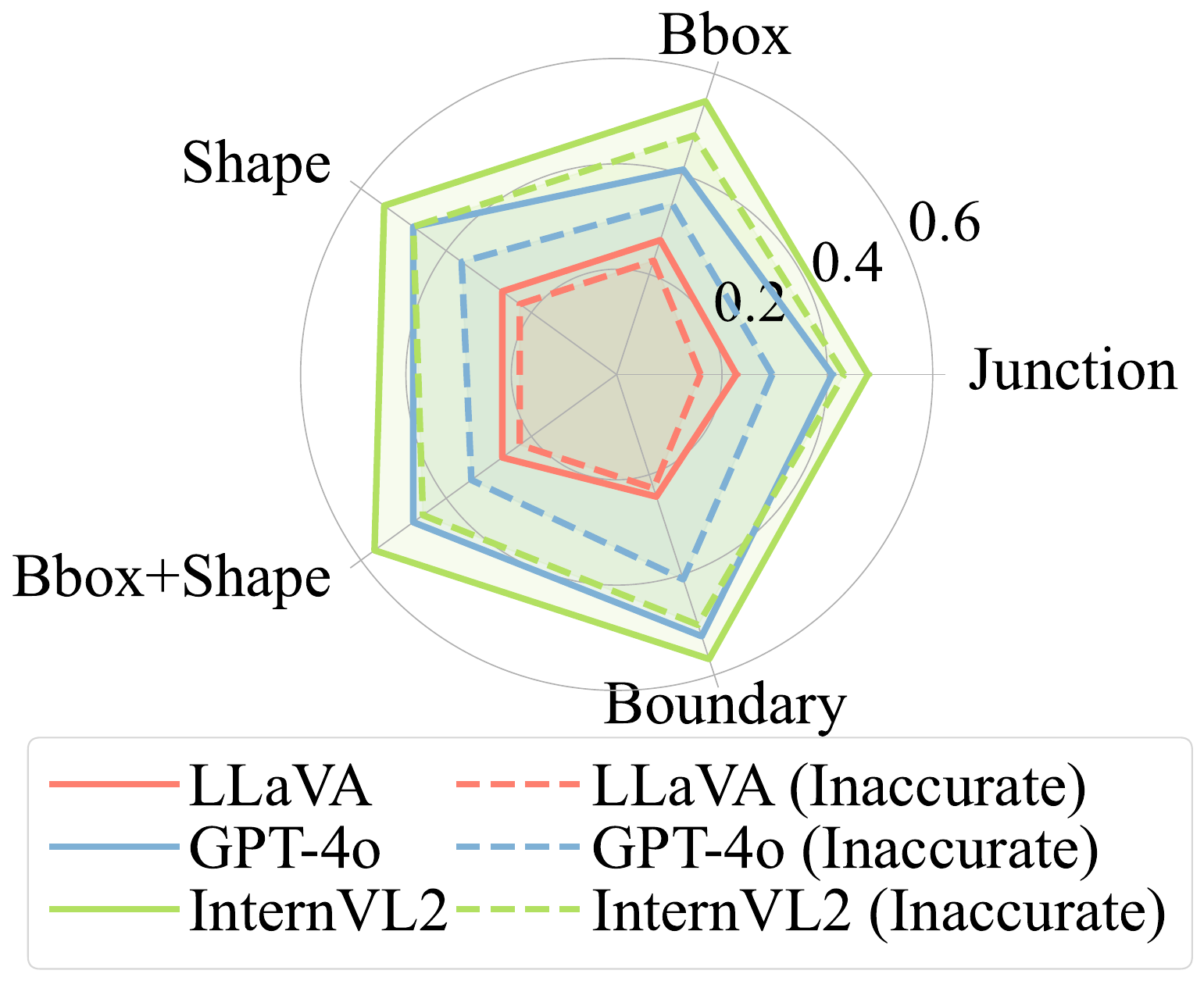}
        \caption{}
        \label{fig:intro_subfig2}
    \end{subfigure}
    \hfill
    \begin{subfigure}[b]{0.32\linewidth}
        \centering
        \includegraphics[width=\linewidth]{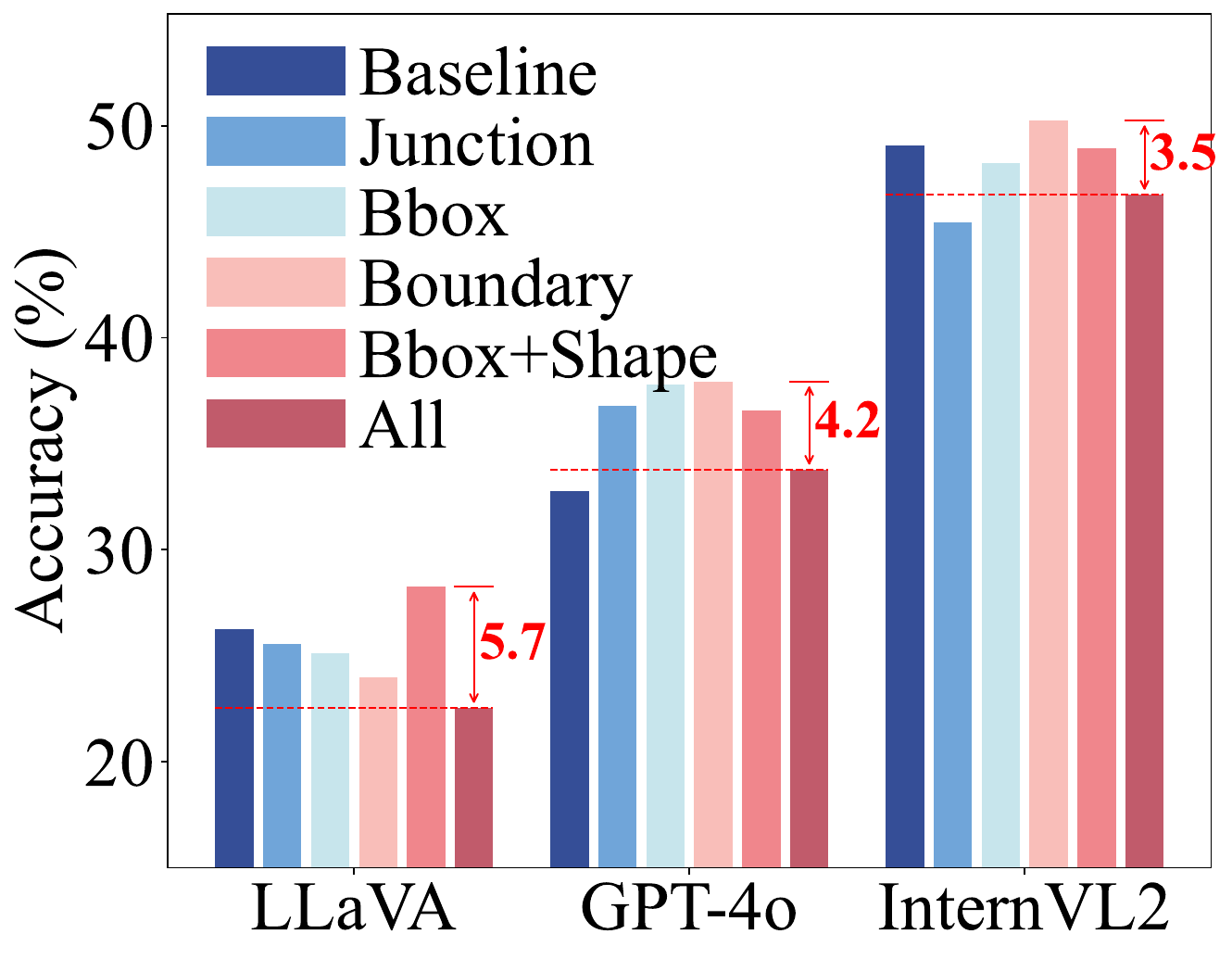}
        \caption{}
        \label{fig:intro_subfig3}
    \end{subfigure}
    
    \caption{Analysis of MLLMs' performance in mathematical visual reasoning tasks from GeoQA test set. GPT-4o misperceived visual information in approximately 70\% of cases involving geometric entities  (Fig.~\ref{fig:intro_subfig1}). Providing optimal geometric information enhances model performance, while redundant visual cues lower top-1 accuracy—even below the baseline achieved with only textual questions. (Fig.~\ref{fig:intro_subfig3}). Model performance is sensitive to the accuracy of visual cues and a significant decrease (~13.6\%) in GPT-4o's top-1 accuracy is observed when provided with inaccurate bounding box locations and shape names (Bbox+Shape)  (Fig.~\ref{fig:intro_subfig2}). }
    \vspace{-0.5cm}
    \label{fig:intro}
\end{figure}

Moreover, despite these efforts, even the most advanced MLLMs still exhibit notable shortcomings in accurately perceiving and grounding basic geometric primitives in mathematical diagrams. Our systematic analysis reveals that visual recognition errors are prevalent and significantly impact the performance of MLLMs on mathematical reasoning tasks. We tasked LLMs with describing geometric entities in meticulously collected 100 images from the Geo170K dataset~\citep{gao2023g}, and then manually reviewed its responses to categorize the correct descriptions and error types. As demonstrated in Fig.~\ref{fig:intro_subfig1}, we observed that GPT-4o misperceived visual information in approximately 70\% of cases involving geometric entities. Correcting these visual perception errors led to a 12\%  overall accuracy improvement on corresponding mathematical questions (refer to Fig. \ref{fig:supp_subfig1} in the Appendix). This finding highlights that misunderstanding visual details remains a critical bottleneck in the mathematical reasoning capabilities of MLLMs.

\begin{minipage}{6.4cm}
To mitigate aforementioned challenges, we propose a novel approach termed SVE-Math (\textbf{S}elective \textbf{V}ision-\textbf{E}nhanced \textbf{Math}ematical MLLM) that diverges from the current trend of scaling up mathematical visual instruction datasets. Instead, we focus on enhancing the fine-grained visual perception capabilities of the model by training an auxiliary visual encoder, \ourgip\ (Geometric-Grounded Language-Image Pre-training), specifically tailored to recognize geometric primitives. Although existing mathematical datasets lack bounding box or pixel-level annotations, the training data generation process is simple yet highly efficient, \eg, through the Matplotlib Python library. Moreover, training protocols for such visual-centric tasks are relatively straightforward compared to those for LLMs. 
\end{minipage}
\begin{wrapfigure}{r}{0.51\linewidth}
\vspace{-7.6cm}
\hspace{0.3cm}
\centering
\includegraphics[trim=3 6 3 5, clip=true, width=1\linewidth]{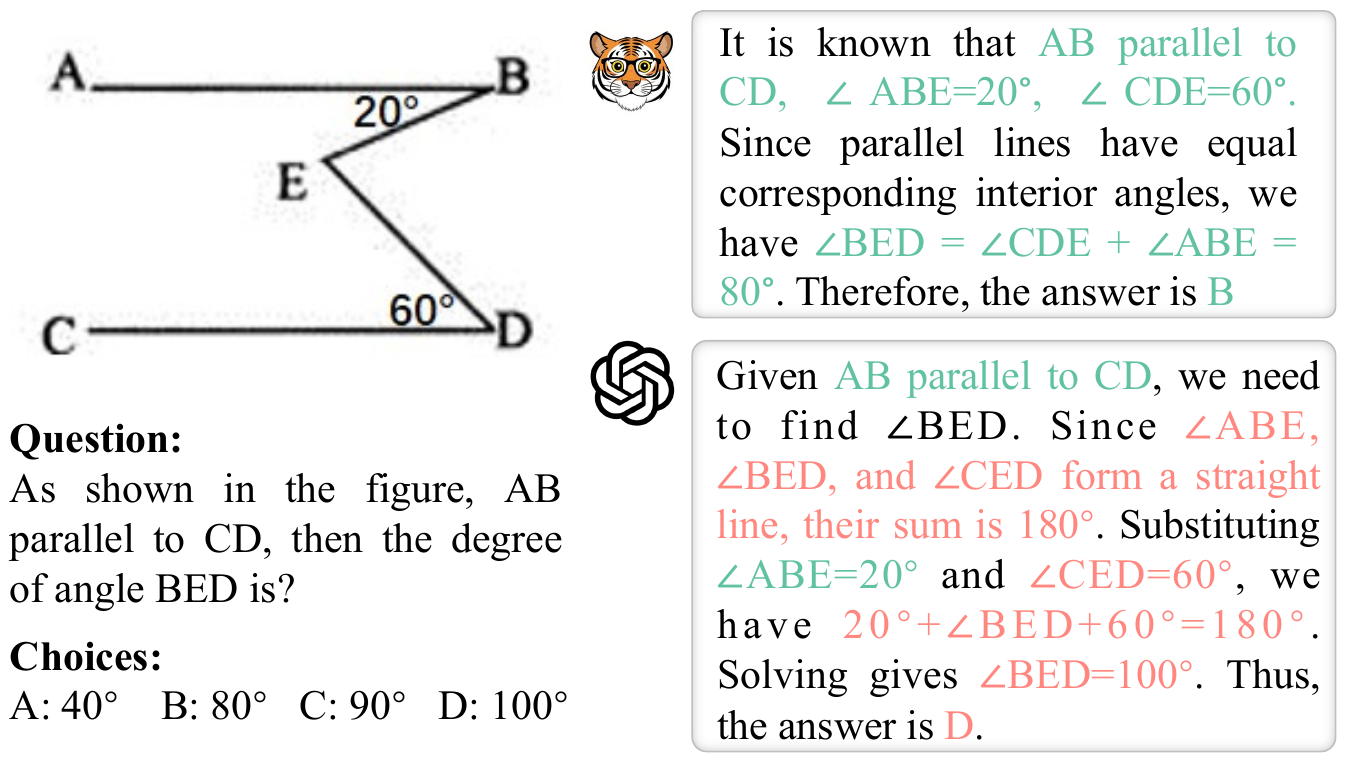}
\vspace{-0.5cm}
\caption*{$\rhd$ GPT-4o (bottom) struggles to accurately perceive mathematical elements, resulting in errors in reasoning about their relationships. By integrating \ourgip, \ourMethod\ (top) effectively grounds geometric elements and their positional relations (\eg, $\angle$CDE), enabling accurate reasoning. See the Appendix for additional examples. }
\label{fig:explain}
\vspace{-1.2cm}
\end{wrapfigure}
By incorporating GeoGLIP into existing MLLMs, we enable the models to \emph{open their eyes} to the essential visual components of mathematical problems before engaging in reasoning.

Our hypothesis and design are inspired by observations as shown in Fig.~\ref{fig:intro_subfig2} and Fig.~\ref{fig:intro_subfig3}. Specifically, instructing MLLMs with fine-grained visual information, such as junction points and object locations, improves top-1 accuracy compared to providing only worded questions. However, providing all visual cues for solving a math question decreases accuracy, \eg, a 4.2\% decrease in GPT-4o's performance. These `apples-to-apples' comparisons highlight that relevance is key—excessive information interferes with problem-solving (see \S ~\ref{case} for a case study). Moreover, their performance is highly sensitive to the accuracy of visual cues. Providing inaccurate instructions, such as randomly generated box locations, significantly decreases performance. Given the inherent uncertainty in detecting geometric primitives by GeoGLIP, our initial approach utilizes global pyramid feature maps, which capture information ranging from geometry-rich to semantic-rich representations. Their contributions are dynamically modulated by the feature router mechanism, resulting in the so-called visual soft prompts.




Our proposed \ourMethod\ has several key advantages. First, by enhancing the visual encoder to perceive geometric primitives, we directly tackle the root cause of geometrical visual recognition errors in mathematical reasoning tasks. Second, \ourMethod\ is efficient and practical, as it does not rely on the creation of large-scale instruction datasets or extensive human annotations. Third, our proposed auxiliary visual encoder and connector can be seamlessly integrated into any existing MLLM, enhancing its performance without modifying the reasoning components of language models.

We evaluate \ourMethod\ on several public mathematical benchmarks, and experimental results demonstrate its superior performance compared to models of the same or even larger sizes. Specifically, our model outperforms other 7B-parameter models and achieves comparable results to advanced 13B-parameter MLLMs, all while using a smaller-scale dataset for visual training (40K) and 60K + 110K for alignment and instruct learning, compared to the large 588K + 834K dataset used in MAVIS~\citep{zhang2024mavis}. These results highlight the effectiveness of our approach and underscore the importance of accurate visual perception in mathematical visual reasoning.
In summary, our contributions are as follows:

\begin{itemize} \setlength\itemsep{0em}
\item We systematically identify and analyze the impact of visual recognition errors on the mathematical reasoning performance of MLLMs, highlighting the importance of accurately perceiving geometric primitives. 
\item We propose \ourMethod, a novel framework that enhances the visual perception capabilities of MLLMs by integrating a geometric-awareness visual encoder trained on small-scale box/pixel-level annotations, reducing the dependency on large-scale instruction datasets.
\item We introduce a connector mechanism with a feature router that dynamically integrates relevant geometric visual information into the language model, improving mathematical reasoning performance without modifying the model's reasoning components.
\item Our \ourgip\ encoder integrates seamlessly with diverse LLM backbones, requiring no architectural changes to their reasoning modules. Extensive experiments demonstrate that \ourMethod\ outperforms existing models of comparable or larger sizes on multiple mathematical reasoning benchmarks. \end{itemize}

\section{Related Work}
\label{sec:relawork}
\noindent \textbf{Multimodal Large Language Models for Mathematics.}
Large Language Models (LLMs) have recently garnered significant attention, with much research focused on text-based mathematical problem-solving, expanding mathematical datasets and utilizing data augmentation ~\citep{yu2023metamath,yue2023mammoth,yue2024mammoth2,luo2023wizardmath}. 
Meanwhile, advancements in vision-language alignment models,  such as CLIP~\citep{Radford2021LearningTV} and BLIP~\citep{li2022blip}, have significantly progressed multimodal tasks, leading to the development of Multimodal Large Language Models (MLLMs)~\citep{bai2023qwen,team2023gemini,ye2023mplugowl,lin2023sphinx,gao2024sphinx,hu2024visual}.  
With the rise of instruction-following LLMs, LLaVA~\citep{liu2024visual} adopts a linear layer to directly project visual tokens into LLMs, while MiniGPT-4~\citep{zhu2023minigpt} resamples visual tokens into fixed-length tokens, reducing the computation cost. 

Building on these advancements, researchers have started to explore visual mathematical problem-solving using MLLMs. Unified frameworks like UniGeo~\citep{Chen2022UniGeoUG}, UniMath~\citep{liang_unimath}, and GeomVerse~\citep{kazemi2023geomverse} expand multimodal mathematical datasets and improve MLLM performance in geometry and diverse tasks. Leveraging current datasets, G-LLaVA~\citep{gao2023g} constructed the Geo170K dataset, enhancing geometric problem-solving and surpassing GPT-4V~\citep{openai2023gpt4v} on MathVista~\citep{Lu2023MathVistaEM} with only 7B parameters. GeoGPT4V~\citep{cai2024geogpt4v} further improved model performance on MathVista and MathVision~\citep{wang2024measuring} by creating a high-quality geometric problem dataset using GPT-4 and GPT-4V. MAVIS~\citep{zhang2024mavis} specializes in mathematical tasks with a three-stage training pipeline including a math-specific vision encoder, while Math-LLaVA~\citep{shi2024math} introduced MathV360K, a large-scale dataset with high-quality images and diverse question-answer pairs to improve multimodal mathematical reasoning. These math-specific MLLMs have shown promising performance across several benchmark datasets~\citep{yue2023mmmu,zhang2024mathverse}.

Despite these advancements, MLLMs still face challenges in multimodal mathematical tasks, particularly due to limitations in visual perception. While CLIP remains a common choice for many mathematical MLLMs and is known to benefit multimodal tasks, its limitations have also been identified. For instance, \citep{tong2024eyes} examines `CLIP-blind pairs', revealing that visually distinct images are often misinterpreted as similar, highlighting systematic shortcomings in CLIP's visual perception. These findings underscore the need for more specialized visual encoding methods tailored to mathematical contexts, as well as more rigorous evaluations of MLLMs' visual capabilities.

\noindent \textbf{Open-Set Object Detection.} Open-set object detection identifies arbitrary classes using existing bounding box annotations and language generalization. Methods like OV-DETR~\citep{zareian2021openvocabularyobjectdetectionusing}, ViLD~\citep{gu2022openvocabularyobjectdetectionvision}, DetCLIP~\citep{yao2022detclipdictionaryenrichedvisualconceptparalleled}, and Grounding DINO~\citep{liu2024groundingdinomarryingdino} integrate language models with detection frameworks to improve category-specific detection. However, these models often struggle with small-scale object detection due to insufficient fine-grained visual understanding. GLIP~\citep{li2022grounded} addresses this limitation by integrating textual information with visual region features early in the pipeline via a language-aware deep fusion mechanism, enhancing region-level embeddings. GLIP improves the detection of smaller objects and demonstrates strong zero-shot capabilities. While GLIP's potential has been explored in various fields~\citep{suris2023vipergpt,peng2023kosmos,li2023gligen}, its application to mathematical reasoning, particularly in precise geometric entity description and fine-grained detail identification in mathematical diagrams, remains largely unexplored. Our work extends these concepts, developing a geometric-grounded language-image pre-training model (\ourgip) tailored for the unique demands of visual mathematical reasoning.

\noindent \textbf{Junction and Boundary Detection.} Junction and boundary detection are crucial in image processing and object recognition~\citep{dollar2006supervised,maire2008using,parida1998junctions}, and can play a pivotal role in mathematical reasoning with geometric diagrams. Junctions represent points where lines intersect, and boundaries delineate object shapes. Traditional methods like Canny edge detection~\citep{canny1986computational} and the Hough Transform~\citep{duda1972use} struggle with complex diagrams and fine-grained details required for accurate mathematical reasoning. Recent deep learning approaches, such as junction detection networks~\citep{huang2018learning}, detect key points by considering surrounding regions. Boundary detection models like Field of Junctions (FoJ)~\citep{verbin2021field} use a bottom-up approach with `generalized M-junctions' to detect contours and junctions. 

\begin{figure}[t]
    \centering
    \vspace{-0.8cm}
    \includegraphics[width=0.95\linewidth]{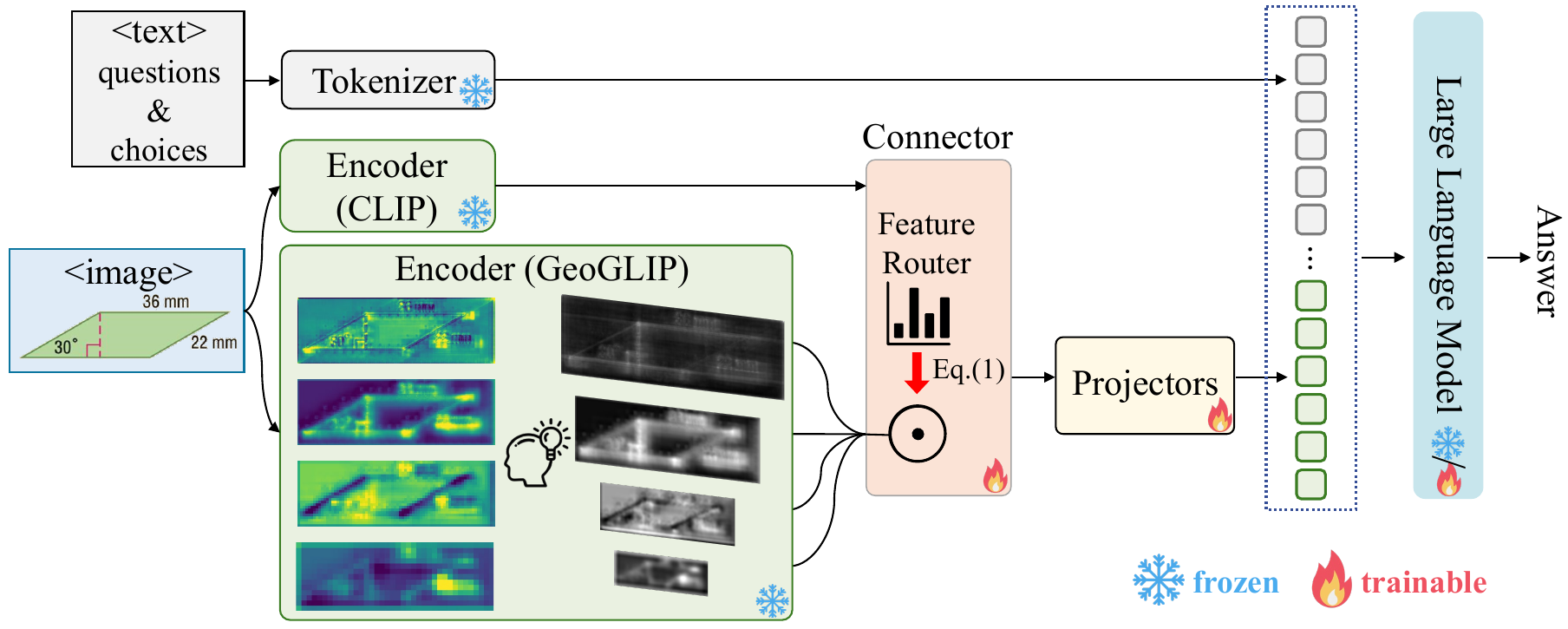}\vspace{-0.1cm}
    \figcaption{The diagram presents the architecture of \ourMethod, highlighting key innovations in the geometric-grounded vision encoder (\ourgip) and the feature router. Fine-grained visual understanding is achieved through a feature pyramid (attention maps displayed on the left), capturing hierarchical visual features ranging from geometry-rich to semantic-rich information. The feature router dynamically adjusts the contribution of these features to generate visual soft prompts. These prompts are then combined with CLIP visual tokens and textual inputs before being fed into the language model (LLM), enabling accurate visual perception and enhanced mathematical reasoning.
    \vspace{-0.6cm}}
    \label{fig:method}
\end{figure}

\section{Methods}
\label{sec:method}
\label{arch}
\ourMethod\ integrates visual understanding of geometric primitives with textual analysis to enhance the model's capability in solving mathematical problems involving visual elements. As illustrated in Fig.~\ref{fig:method},  our pipeline builds upon the LLaVA-1.5~\citep{liu2023llava} architecture (refer to \S \ref{sec:bg}), introducing key innovations in the \ourgip\ and visual feature connector. Feature maps from different layers of the \ourgip\ encoder are processed through the connector, where a feature router optimally integrates the feature pyramid into visual soft prompts by leveraging geometric information. These visual prompts are then fused with CLIP vision tokens, either along the sequence dimension or the channel dimension, and aligned with text embeddings via projection layers for visual understanding. Since channel-wise fusion offers better computational efficiency and comparable performance to sequence-based fusion in our experiments, we set channel-wise fusion as the default approach.

\subsection{Geometric-Grounded Language-Image Pre-trainin}
\label{MAGLIP}
Our proposed \ourgip\ extends GLIP~\citep{li2022grounded} to perform shape grounding, boundary and junction detection tasks with no human annotations. The architecture of \ourgip\ is shown in Fig.~\ref{fig:glip} of the Appendix. For shape grounding, we follow the same pipeline structure as the original GLIP model for bounding box detection (refer to \S \ref{sec:bg} for pipeline details) but train it on the mathematical domain. Unlike the grounding task, which prioritizes semantic-rich visual information for localizing objects based on text inputs, boundary and junction detection require finer visual details. In general, feature pyramids encode information at different levels: higher-resolution features capture more geometric details, while lower-resolution features capture more semantic information. We employ a cross-resolution mixture to inject low-resolution features into high-resolution features, thereby improving visual understanding. Training details are provided in \S~\ref{train}, and the training datasets are discussed in \S~\ref{syndata}. Visualization results can be seen in Figures~\ref{fig:devis} and~\ref{fig:devis2} of the Appendix.

\noindent \textbf{Boundary and junction detection.} 
GLIP-T utilizes Swin-Tiny as its backbone, producing a five-level feature pyramid $\{F_{\text{geo}}^{i}\}_{i \in \{1,2,3,4,5\}}$, where each level's resolution is progressively downscaled by a factor of 2. To enrich the high-resolution features with semantic information, we first pass the high-resolution tensor $F_{\text{geo}}^{2}$ (as the Query) and the low-resolution tensor $F_{\text{geo}}^{4}$ (as the Key and Value) to a Multi-Head Self Attention (MHSA) module. The resulting feature maps are upsampled by a factor of 2 and element-wise added to $F_{\text{geo}}^{1}$, producing $F_{\text{geo}}^{1^*}$. The rationale behind this design is to fully integrate the hierarchical object concepts at various scales produced by the downsampling layers with the high-resolution spatial information encoded by the initial embedding layer. Taking $F_{\text{geo}}^{1^*}$ as input, we then adopt two decoders for boundary and junction detection (see Fig. \ref{fig:detail}).


The boundary decoder consists of two successive perception blocks, each comprising an upsampling operation using nearest-neighbor interpolation, followed by a $3 \times 3$ convolution (Conv2d), batch normalization (BN2d), and ReLU activation. The final output is resized to the original image resolution using bilinear upsampling.

A junction represents the intersection of lines, determined by the intersection coordinates and the orientations of the lines. Accordingly, our junction decoder has two branches. The first branch estimates the confidence of a junction falling within each grid cell of the original image (using a $60 \times 60$ grid) and its relative position to the cell's center coordinates. The second branch predicts the orientations of the intersecting lines and their confidence in falling into one of 15 evenly spaced bins within each grid cell, where each bin covers 24 degrees, ensuring the full 360-degree range is divided evenly (15 bins $\times$ 24 degrees = 360 degrees). In the junction decoder, the input $F_{\text{geo}}^{1^*}$ is first processed through a perception block, where it is upsampled to a resolution of $60 \times 60$. Then, two separate Conv2D units predict the cell confidence and location, with output sizes of $60 \times 60 \times 1$ and $60 \times 60 \times 2$, respectively. Additionally, two other Conv2D units predict the bin confidence and orientation, both producing outputs of $60 \times 60 \times 15$. For further details, refer to training step 1 in \S \ref{train} and the illustration in Fig.~\ref{fig:detail} in the Appendix.


\subsection{Connector Design}
\label{connector}

Recall our hypothesis that selecting key visual cues enhances mathematical visual problem-solving, while redundant information can hinder it. To manage the contribution of each feature and enhance the model's capacity, we propose a dynamic feature router $R$. The router $R$ is implemented as a simple Multi-Layer Perceptron (MLP) that takes as input the concatenation of the spatially averaged pooled feature maps from each level of \ourgip\ ($\boldsymbol{\bar}F_{\text{geo}}^i \in \mathbb{R}^{1 \times 256}$)  and the CLIP feature map ($\boldsymbol{\bar}F_{\text{clip}} \in \mathbb{R}^{1 \times 1,024}$). It calculates the routing weights per feature ($\{\boldsymbol{w}^i\}_{i\in \{1,2,3,4\}} \in \mathbb{R}^{1 \times 4}$), functioning as a soft router~\citep{softmixturesexperts}.  Alternative types of routers, such as sparse routers and constant routers, are also discussed in Sec.~\ref{sec:exp}. The soft router's process is defined as:
\begin{align} \widehat{F}^i_{\text{geo}} &= \mathbf{w}^i \cdot \text{MLP}\left(\mathcal{G}\left(F_{\text{geo}}^i\right)\right), \ \mathbf{w}^i = \text{SoftMax}\left([\boldsymbol{\bar}F_{\text{geo}}^i, \boldsymbol{\bar}F_{\text{clip}}]\right), \end{align}
where $F_{\text{geo}}^i$ is resized by $\mathcal{G}$ to match the spatial dimensions of $F_{\text{clip}}$, and subsequently processed by a multi-layer perceptron (MLP) to align channel dimensions. The scalar routing weights $\mathbf{w}^i$ are computed via a learned router $R$, which operates on the concatenation of normalized global feature maps $[\overline{F}_{\text{geo}}^i, \overline{F}_{\text{clip}}]$. 


The final $\widehat{F}_{\text{geo}}$ is generated using one of two fusion strategies depending on the fusion strategy with $F_{\text{clip}}$: 
(1) element-wise addition of the weighted features, 
$\widehat{F}_{\text{geo}} = \sum_{i=1}^{4} \widehat{F}^i_{\text{geo}}$, 
where the weights $\mathbf{w}^i$ are normalized via the SoftMax function (i.e., $\sum_{i=1}^{4} \mathbf{w}^i = 1$); or 
(2) channel-wise concatenation of the weighted features, 
where the weights are processed through a Sigmoid function. depending on the fusion strategy with $F_{\text{clip}}$.




Next, we explore strategies for fusing the soft prompts $\widehat{F}_{\text{geo}}$ with $F_{\text{clip}}$, either sequence-wise or channel-wise. In the sequence-wise method, additional tokens are added after the CLIP tokens, 
\begin{wrapfigure}{r}{0.33\textwidth} 
    \centering
     \vspace{-0.3cm}
  \includegraphics[width=0.32\textwidth, height=0.18\textwidth]{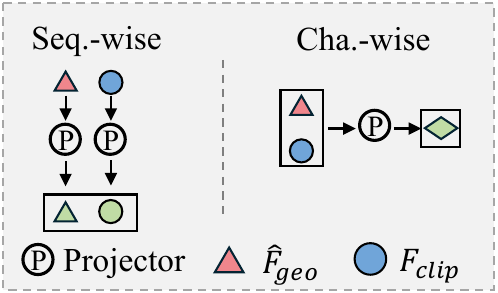}
\end{wrapfigure}
extending the sequence length. In contrast, channel-wise fusion combines all visual tokens along the channel dimension, maintaining the same sequence length. To enable the subsequent LLM to understand these visual components,  the fused visual tokens are then fed into projection layers, which project the visual modality into the LLM's embedding space. Following the LLaVa-1.5 approach, we employ highly effective MLP projectors (linear layer + GELU + linear layer, \aka, \texttt{mlp2x\_gelu}) for this task. In the sequence-wise approach, two separate projectors are applied for CLIP and soft prompts, respectively. For example, the projection matrices for the two linear layers, per projector, $\boldsymbol{\Phi_1}$ and $\boldsymbol{\Phi_2}$, have sizes of $1,024 \times 4,096$ and $4,096 \times 4,096$, where $4,096$ corresponds to the text embedding dimension. In the channel-wise approach, a single projector ($\boldsymbol{\Phi_1} \in \mathbb{R}^{5,120 \times 4,096}$ and $\boldsymbol{\Phi_2} \in \mathbb{R}^{4,096 \times 4,096}$) is used to process the combined visual tokens. 


\begin{figure}[t]
    \centering
    \vspace{-1cm}
    \includegraphics[width=0.9\linewidth]{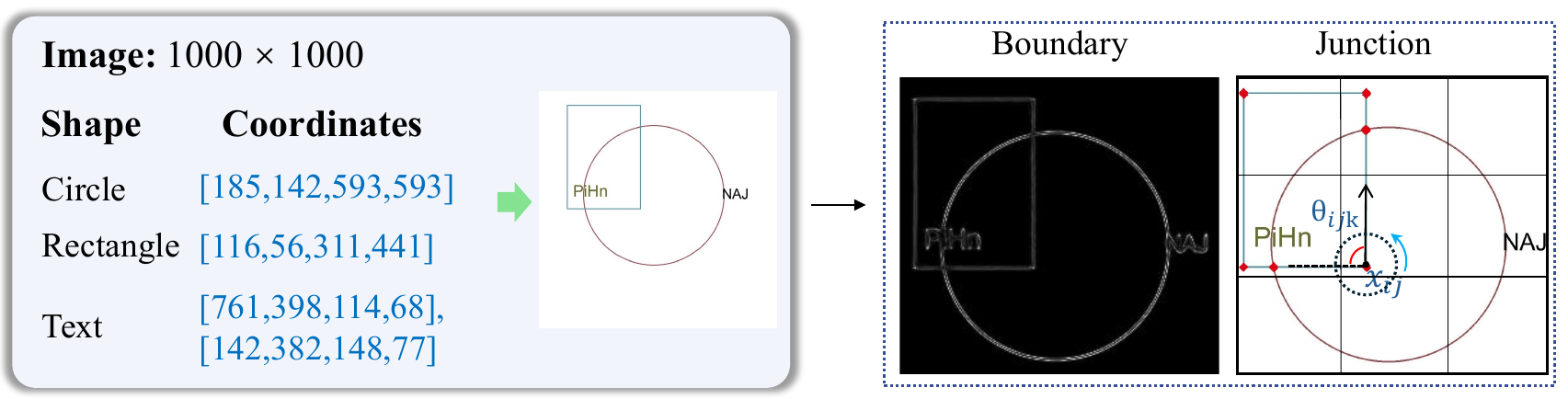}\vspace{-0.01cm}
    \figcaption{Process for generating synthetic data with box- and pixel-level annotations for training our \ourgip\ visual encoder. Each image contains geometric objects such as circles, rectangles, and alphanumeric text (`Text') with random strings of length 1 to 10 placed alongside geometric shapes. Refer to Fig.~\ref{supp:prorgam} in the Appendix for a detailed flowchart of the generation pipeline.\vspace{-0.4cm}}
    \label{intro:flow}
\end{figure}

\subsection{Training samples for visual-centric \ourgip}
To enable \ourgip\ to perceive fine-grained mathematical elements, we supervise its training using datasets with box- and pixel-level annotations. The model is trained with a classical detection loss $\mathcal{L}_{det}$ (Eq. \ref{equ:loss_det}), a junction loss $\mathcal{L}_{junc}$ (Eq. \ref{equ:loss_jun}), and a boundary loss $\mathcal{L}_{bodr}$ (the $\ell_2$ loss between predicted heatmap values and ground truth values). The detection loss $\mathcal{L}_{det}$ is applied to the shape grounding task, using synthetic images and FigureQA~\citep{figuredataset} training data annotated with bounding boxes and shape names (left panel of Fig.~\ref{intro:flow}). These annotations are stored in a COCO-style JSON file for seamless integration with standard GLIP. See \S \ref{syndata} for details on the synthetic data engine and dataset statistics (Figures \ref{fig:supp_subfig2} and \ref{fig:supp_subfig3}). 

For boundary and junction detection tasks, we leveraged off-the-shelf models~\citep{huang2018learning,verbin2021field} to extract junctions and boundaries as ground truth. In addition to our synthetic samples, we incorporated the public dataset Geo170K~\citep{chen2021geoqa} and generated the corresponding ground truth. Specifically, junction labels include intersection coordinates and line orientations. As noted, each grid cell and bin are responsible for predicting the coordinates and the orientations, and we have 60 $\times$ 60 cells\&15 bins per cell.  The labels are formatted as $JP_{ij} = (x_{ij}, c_{ij}, \{\theta_{ijk}, c^{\theta}_{ijk}\}_{k=1}^{K})$, where $x_{ij}$ denotes the junction center coordinates, $c_{ij} \in \{0,1\}$ indicates the presence of a junction, $\theta_{ijk}$ is the angle of the $k$-th bin, and $c^{\theta}_{ijk} \in \{0,1\}$ is the indicator for that bin (right panel of  Fig. \ref{intro:flow}). 



\begin{table*}[!t]
\vspace{-0.9cm}
\small
\centering\caption{\textbf{Results on testmini set of MathVerse} with the accuracy metric. The highest results for \colorbox{backred!50}{closed-source} and \colorbox{backblue!75}{open-source} MLLMs are highlighted in red and blue respectively. }
\begin{adjustbox}{width=1\linewidth}
    \begin{tabular}{l|c|C{0.9cm}|C{1.25cm}|C{0.9cm}|C{1.1cm}|C{1.25cm}|C{0.9cm}}
    \toprule
    \multirow{3}*{\makecell*[l]{\large Model}}  &\multirow{3}*{\makecell*[l]{Base\\LLM}} &\multicolumn{1}{c|}{\makecell*[c]{All}}
    &\makecell*[c]{\shortstack{Text\\\vspace*{0.2pt}\\Dominant\\\vspace*{0.1pt}}}
    &\makecell*[c]{\shortstack{Text\\\vspace*{0.2pt}\\Lite\\\vspace*{0.1pt}}}
    &\makecell*[c]{\shortstack{Vision\\\vspace*{0.2pt}\\Intensive\\\vspace*{0.1pt}}}
    &\makecell*[c]{\shortstack{Vision\\\vspace*{0.2pt}\\Dominant\\\vspace*{0.1pt}}}
    &\makecell*[c]{\shortstack{Vision\\\vspace*{0.2pt}\\Only\\\vspace*{0.1pt}}}\\
    \cmidrule{3-8}
    &&Acc &Acc &Acc &Acc &Acc &Acc \\
    \midrule
    \multicolumn{8}{c}{\textit{Baselines}}\\
    \cmidrule{1-8}
    Random Chance &-&12.4&12.4&12.4&12.4&12.4&12.4  \\
    Human  &-&67.7&71.2&70.9&61.4&68.3&66.7 \\
    \cmidrule{1-8}
    \multicolumn{8}{c}{\textit{LLMs}}\\
    \cmidrule{1-8}
    ChatGPT~\citep{ouyang2022training} &-&26.1&33.3&18.9&-&-&-\\
    GPT-4~\citep{OpenAI2023GPT4TR} &-&33.6&46.5&46.5&-&-&-\\
    \cmidrule{1-8}
    \multicolumn{8}{c}{\textit{Closed-source MLLMs}}\\
    \cmidrule{1-8}
    Qwen-VL-Plus~\citep{bai2023qwen} &-&11.8&15.7&11.1&9.0&13.0&10.0\\
    Gemini-Pro~\citep{team2023gemini} &-&23.5&26.3&23.5&23.0&22.3&22.2 \\
    Qwen-VL-Max~\citep{bai2023qwen} &-&25.3&30.7&26.1&24.1&24.1&21.4   \\
    GPT-4V~\citep{openai2023gpt4v} &- &\colorbox{backred!50}{39.4}&\colorbox{backred!50}{54.7}&\colorbox{backred!50}{41.4}&\colorbox{backred!50}{34.9}&\colorbox{backred!50}{34.4}&\colorbox{backred!50}{31.6} \\
    \cmidrule{1-8}
    \multicolumn{8}{c}{\textit{Open-source MLLMs}}\\
    \cmidrule{1-8}
    LLaMA-Adapter V2~\citep{gao2023llamaadapterv2}&\tiny LLaMA-7B~\citep{touvron2023llama}&5.7&6.2&5.9&6.1&4.2&6.1 \\
    ImageBind-LLM~\citep{han2023imagebind}&\tiny LLaMA-7B&9.2&11.4&11.3&8.9&11.2&3.4  \\
    mPLUG-Owl2~\citep{ye2023mplugowl2} &\tiny LLaMA-7B&5.9&6.6&6.3&6.3&5.6&4.9\\
     SPHINX-Plus~\citep{gao2024sphinx} &\tiny LLaMA2-13B &12.2&13.9&11.6&11.6&13.5&10.4 \\
    SPHINX-MoE~\citep{gao2024sphinx}&\tiny Mixtral-8$\times$7B~\citep{albert24mixtral}&15.0&22.2&16.4&14.8&12.6&9.1  \\
    G-LLaVA~\citep{gao2023g}&\tiny LLaMA2-7B&16.6&20.9&20.7&17.2&14.6&9.4 \\
    InternLM-XC2.~\citep{dong2024internlm}&\tiny InternLM2-7B~\citep{cai2024internlm2}&16.5&22.3&17.0&15.7&16.4&11.0 \\
    LLaVA-1.5~\citep{liu2023improvedllava} &\tiny Vicuna-13B&7.6&8.8&7.6&7.4&7.4&6.9\\
    ShareGPT4V~\citep{Chen2023ShareGPT4VIL} &\tiny Vicuna-13B&13.1&16.2&16.2&15.5&13.8&3.7 \\
    Math-LLaVA~\citep{shi2024math}&\tiny Vicuna-13B&19.0&21.2&19.8&20.2&17.6&16.4 \\
    LLaVA-NeXT~\citep{li2024llavanext-strong} &\tiny LLaMA3-8B~\citep{qwen1.5}&19.3&24.9&20.9&20.8&16.1&13.8\\
     \bf{\ourMethod-7B} &\tiny LLaMA2-7B  &{21.2}&{26.4}&{23.2}&{22.9}&{18.0}&{15.4} \\
    \bf{\ourMethod-Deepseek-7B} &\tiny Deepseek-math-7B~\citep{deepseek2023math}&{24.3}&{31.1}&{26.9}&{25.6}&{19.3}&\colorbox{backblue!75}{17.5} \\
     \bf{\ourMethod-Qwen2.5-7B} &\tiny Qwen-math-7B~\citep{qwen2023math}&\colorbox{backblue!75}{31.4}&\colorbox{backblue!75}{37.6}&\colorbox{backblue!75}{36.8}&\colorbox{backblue!75}{34.9}&\colorbox{backblue!75}{31.5}&{16.0} \\
    \bottomrule
    \end{tabular}
\end{adjustbox}
\label{t1}
\vspace{-0.3cm}
\end{table*}

\begin{table*}[!t]
\vspace{-0.3cm}
\small
\centering\caption{\textbf{Results on testmini set of MathVista} with the accuracy metric. The highest results for \colorbox{backred!50}{closed-source} and \colorbox{backblue!75}{open-source} MLLMs are highlighted.} 
\begin{adjustbox}{width=1\linewidth}
    \begin{tabular}{l|c|C{0.9cm}|C{1.25cm}|C{0.9cm}|C{1.1cm}|C{1.25cm}|C{0.9cm}}
    \toprule
    \multirow{3}*{\makecell*[l]{\large Model}}  &\multirow{3}*{\makecell*[l]{Base\\LLM}} &\multicolumn{1}{c|}{\makecell*[c]{All}}
    &\makecell*[c]{FQA}
    &\makecell*[c]{GPS}
    &\makecell*[c]{MWP}
    &\makecell*[c]{TQA}
    &\makecell*[c]{VQA}\\
    \cmidrule{3-8}
    &&Acc &Acc &Acc &Acc &Acc &Acc \\
    \midrule
    \multicolumn{8}{c}{\textit{Baselines}}\\
    \cmidrule{1-8}
    Random Chance &-&17.9&18.2&21.6& 3.8 &19.6& 26.3  \\
    Human  &-&60.3&59.7&48.4&73.0&63.2&55.9 \\
    \cmidrule{1-8}
    \multicolumn{8}{c}{\textit{Closed-source MLLMs}}\\
    \cmidrule{1-8}
    Qwen-VL-Plus~\citep{bai2023qwen}&- &{43.3}&\colorbox{backred!50}{54.6}&{33.5}&31.2&48.1&\colorbox{backred!50}{51.4} \\
     GPT-4V~\citep{openai2023gpt4v} &-&\colorbox{backred!50}{49.9}&43.1&\colorbox{backred!50}{50.5}& \colorbox{backred!50}{57.5}& \colorbox{backred!50}{65.2}&38.0\\
    \cmidrule{1-8}
    \multicolumn{8}{c}{\textit{Open-source MLLMs}}\\
    \cmidrule{1-8}
    mPLUG-Owl2~\citep{ye2023mplugowl2} &\tiny LLaMA-7B&22.2&22.7&23.6 &10.2 &27.2&27.9\\
     MiniGPT-v2~\citep{chen2023minigpt}&\tiny LLaMA2-7B~\citep{touvron2023llama2} &23.1&18.6&26.0 &13.4 &30.4&30.2 \\
    G-LLaVA~\citep{gao2023g}&\tiny LLaMA2-7B&25.1&19.1&48.7&3.6&25.0&28.7 \\
    LLaVA-1.5~\citep{liu2023improvedllava} &\tiny Vicuna-13B&27.7&23.8&22.7 &18.9&43.0&30.2\\
    SPHINX-Plus~\citep{gao2024sphinx} &\tiny LLaMA2-13B &{36.7}&\colorbox{backblue!75}{54.6}&16.4&23.1&{41.8}&\colorbox{backblue!75}{43.0} \\
    \bf{\ourMethod-7B} &\tiny LLaMA2-7B&{37.4}&{31.9}&{53.9}&{29.0}&{41.4}&{30.8} \\
     \bf{\ourMethod-Deepseek-7B} &\tiny Deepseek-math-7B~\citep{deepseek2023math}&{48.7}&{37.6}&{63.0}&{48.7}&{48.1}&{35.8} \\
      \bf{\ourMethod-Qwen2.5-7B} &\tiny Qwen-math-7B~\citep{qwen2023math}&\colorbox{backblue!75}{50.4}&{38.7}&\colorbox{backblue!75}{67.3}&\colorbox{backblue!75}{58.1}&\colorbox{backblue!75}{51.2}&{31.8} \\
    \bottomrule
    \end{tabular}
\end{adjustbox}
\label{t2}
\vspace{-0.5cm}
\end{table*}

\section{Experiments}
\label{sec:exp}
\subsection{Experimental Setup}
\noindent \textbf{Implementation Details.} 
Our work follows a structured three-stage training pipeline, including multi-task visual perception training for \ourgip, visual-language alignment, and mathematical instruction tuning for MLLMs (refer to \S \ref{train} for details). We fine-tuned our \ourgip\ model using GLIP-T~\citep{li2022grounded} as the pre-trained model, leveraging a combined dataset of 10,000 synthetic images, 20,672 images from FigureQA, and 9,426 images from the Geo170K training set. Training is conducted on 8 A100 GPUs with a batch size of 32. The base learning rate is set to $1 \times 10^{-5}$ for the language backbone and $1 \times 10^{-4}$ for all other parameters, and it is decreased by a factor of 0.1 at 67\% and 89\% of the total training steps. We employ the same data augmentation strategies as GLIP, including random horizontal flipping and aspect ratio-preserving resizing with a minimum size of 800 pixels. 

For multi-modal training, we freeze the \ourgip\ encoder. In Stage 2, we train only the projection layers to align diagram-language pairs. In Stage 3, we unfreeze both the projection layer and the LLM to perform comprehensive instruction-following tuning. Our \ourgip\ together with a pretrained vision transformer (CLIP ViT-L)\citep{Radford2021LearningTV} are integrated into the language models LLAMA-2\citep{touvron2023llama2}, DeepSeek-Math-7B-Instruct~\cite{deepseek2023math}, and Qwen2.5-Math-7B-Instruct~\cite{qwen2023math}, respectively.  Images are padded to squares and resized to $448 \times 448$ pixels with a white background for processing by CLIP, and to $1000 \times 1000$ pixels for processing by \ourgip. We train \ourMethod\ for one epoch for cross-modal alignment and two epochs for instruction tuning on the Geo170K\citep{gao2023g} dataset, evaluating the model on GeoQA~\citep{gao2023g}. Follow the Math-LLaVA-13B~\citep{shi2024math}, we train our model on MathV360k ~\citep{shi2024math} using a batch size of 16 for one epoch with an initial learning rate of $3 \times 10^{-5}$, evaluating on MathVista~\citep{Lu2023MathVistaEM} and the minitest set of MathVerse~\citep{zhang2024mathverse}. 
 
\noindent \textbf{Evaluation Benchmarks.}
We assess our \ourMethod\ using three well-established public mathematical benchmarks, MathVerse~\citep{zhang2024mathverse}, GeoQA~\citep{gao2023g}, and MathVista~\citep{Lu2023MathVistaEM}). MathVerse focuses on assessing multi-modal mathematical problem-solving with a combination of text and diagram-based reasoning tasks. GeoQA emphasizes geometric reasoning, where the model must interpret geometric shapes and solve related questions. MathVista includes a diverse set of mathematical and visual tasks (\eg, IQTest, PaperQA, and IconQA), providing a comprehensive evaluation across various reasoning and problem-solving domains.

\noindent \textbf{Evaluation Metrics.} We adopt top-1 accuracy to evaluate our model on these benchmarks. Our evaluation process follows the protocols defined by the respective datasets, where LLMs are used to extract predicted answers from the model's responses. Accuracy is determined by comparing these predicted answers against the corresponding ground truths.
\subsection{Main Results}
Table~\ref{t1} presents the comparison results on the testmini set of MathVerse, where \ourMethod-7B outperforms all models using LLaMA2-7B as the base LLM by a significant margin (a 5.5\% increase) and achieves comparable top-1 accuracy to the most powerful open-source LLaVA-NeXT~\citep{liu2024llavanext} with 8B size (19.3\% \vs 21.2\%). When using Qwen2.5-Math-7B-Instruct~\citep{deepseek2023math} as the base LLM, our model's performance further increases by an additional +10.2\%. Notably, even on the challenging MathVista benchmark, our model outperforms the advanced SPHINX-Plus-13B~\citep{gao2024sphinx}, and is compatible with close-sourced GPT-4V~\citep{openai2023gpt4v}, as shown in Table~\ref{t2}. This superior performance underscores the importance of fine-grained visual perception in enhancing the mathematical reasoning capabilities of MLLMs.

Tables~\ref{t8} and~\ref{t9} present our model's performance on plane geometry and function analysis tasks, respectively. Compared to the second-best model, MAVIS~\citep{zhang2024mavis}, which is trained on an 8$\times$ larger mathematical visual instruction dataset, \ourMethod\ with LLaMA2-7B as LLM demonstrates better reasoning and generalization capabilities. Furthermore, when GeoGLIP is integrated with reasoning-optimized LLMs (\eg, Qwen2.5-Math-7B-Instruct), our model achieves an additional performance boost exceeding 10\%. Constructing large instruction datasets for training MLLMs is labor-intensive and costly, whereas synthetic datasets for training traditional visual-only tasks offer a more efficient solution. This positions our method, without altering the reasoning process, as a promising and complementary solution for enhancing mathematical visual reasoning tasks.

\begin{figure*}[t]
\centering
\begin{minipage}[c]{0.45\textwidth}
\small
\centering
\tabcaption{Comparison of geometric numerical answer accuracies (\%) on \textbf{GeoQA}.}
\vspace{0.1cm}
\label{t8}
\centering
\begin{adjustbox}{width=0.94\linewidth}
\centering
\begin{tabular}{l|c}
\toprule
\textbf{Model} & \textbf{Accuracy (\%)} \\ 
\midrule
Random Chance &25.0  \\ 
Frequent Guesses &32.1  \\ 
\midrule
\multicolumn{2}{c}{\textit{Top-10 Accuracy}} \\
NGS~\citep{chen-etal-2021-geoqa} & 56.9\\
DPE-GPS~\citep{cao2022augmented} & 62.7\\
SCA-GPS~\citep{sca_gps} & 64.1\\
\midrule
\multicolumn{2}{c}{\textit{Top-1 Accuracy}} \\
Geoformer~\citep{chen2022unigeo} & 46.8\\
UniMath~\citep{liang_unimath} & 50.0\\
G-LLaVA~\citep{gao2023g}  &  64.2 \\ 
MAVIS-7B~\citep{zhang2024mavis}&  {{66.7}}\\ 
\bf{\ourMethod-7B}& {{67.0}}\\ 
\bf{\ourMethod-Deepseek-7B}& {{72.8}}\\ 
\bf{\ourMethod-Qwen2.5-7B}&  \colorbox{backblue!75}{{79.6}}\\ 
\bottomrule
\end{tabular}
\end{adjustbox}
\end{minipage}
\quad
\begin{minipage}[c]{0.48\textwidth}
\small
\centering
\tabcaption{Comparison of model performance on  \textbf{FunctionQA of MathVista.}}
\vspace{0.1cm}
\label{t9}
\begin{adjustbox}{width=0.95\linewidth}
\centering
\begin{tabular}{l|c}
\toprule
\textbf{Model} & \textbf{Accuracy (\%)} \\ \midrule
Random Chance &22.5  \\ 
\midrule
\multicolumn{2}{c}{\textit{Closed-source MLLMs}} \\
CoT GPT-4~\citep{OpenAI2023GPT4TR} & 35.0\\
PoT GPT-4~\citep{OpenAI2023GPT4TR} & 37.0 \\
Multimodal Bard~\citep{google2023bard} & 45.5\\
GPT-4V~\citep{openai2023gpt4v} & \colorbox{backred!50}{69.5}\\
\midrule
\multicolumn{2}{c}{\textit{Open-source MLLMs}} \\
LLaVA~\citep{liu2023llava} & 20.5 \\
LLaMA-Adapter V2~\citep{gao2023llamaadapterv2} & 32.0\\
LLaVA-NeXT~\citep{liu2024llavanext} & 33.7\\
SPHINX-MoE~\citep{gao2024sphinx} & 34.6\\

{MAVIS-7B}~\citep{zhang2024mavis}& {{40.3}}\\ 
\bf{\ourMethod-7B}&  {{40.5}}\\ 
\bf{\ourMethod-Deepseek-7B}&  {{45.1}}\\ 
\bf{\ourMethod-Qwen2.5-7B}&  \colorbox{backblue!75}{{53.3}}\\ 
\bottomrule
\end{tabular}
 \end{adjustbox}
\end{minipage}
\vspace{-0.3cm}
\end{figure*}

\subsection{Ablation Analysis}
\label{exp:abla}
\noindent \textbf{Effect of plug-in geometry-aware visual prompts.} We design \ourgip\, a lightweight, geometry-aware visual model with multitask learning capabilities, including shape grounding, junction detection, and boundary detection. GeoGLIP, combined with a soft router, integrates seamlessly with diverse LLM backbones without requiring modifications to their reasoning components. Despite adding less than a 50MB increase in parameter size and only a 0.24s increase in inference time per image (refer to Table \ref{tab:parameter_comparison} for details), and without relying on additional mathematical instruction datasets, our approach achieves an 8–12\% improvement in top-1 accuracy compared to the baseline (G-LLaVA, using LLaMA2-7B as the base LLM), as shown in Tables \ref{t1}-\ref{t8}. Further ablation studies in Table \ref{tab:model_comparison} reveal a significant drop in Top-1 accuracy on the challenging MathVista testmini set when GeoGLIP is removed (SVE-Math(-)), highlighting the generalizability and effectiveness of our approach.  These results indicate that SVE-Math complements reasoning-focused approaches by bridging the gap in visual perception—an area less emphasized in current state-of-the-art designs.

\noindent \textbf{Effect of cross-resolution mixture.}
We designed four additional variants to demonstrate the effectiveness of our cross-resolution mixture approach. Recall that we have five feature levels $\{F_{\text{geo}}^{i}\}_{i \in \{1,2,3,4,5\}}$ with different resolutions, each with different resolutions, ranging from geometric-rich to semantic-rich information. The cross-resolution mixture aims to generate the input $F_{\text{geo}}^{1^*}$ for the boundary and junction decoders, with the expectation that $F_{\text{geo}}^{1^*}$ captures more informative visual information to benefit boundary and junction detection tasks. 

Using boundary detection as an example, we first used the semantic-rich $F_{\text{geo}}^5$ as input to the boundary decoder. As shown in Fig.~\ref{bun1}, the decoder fails to generate clear boundaries, resulting in a blurred output. Next, we used the geometric-rich $F_{\text{geo}}^1$, which performs better (Fig.~\ref{bun2}), showing some visible boundaries. To further enhance the results, we applied a cross-resolution attention mechanism (classic Multi-Head Self-Attention, MHSA) between $F_{\text{geo}}^{2}$ and $F_{\text{geo}}^{4}$, improving boundary detection as seen in Fig.~\ref{bun4}. Since boundary detection benefits from geometric-rich information, we upsampled the cross-correlated features by a factor of 2 and added them element-wise with $F_{\text{geo}}^{1}$, producing the best visualization results, especially for finer details (Fig.~\ref{bun5}). Finally, to assess the importance of cross-resolution attention, we replaced it with element-wise addition. As expected, the boundaries became blurred (Fig.~\ref{bun3}) due to the reduced receptive field. Replacing addition with the attention mechanism yields similar boundary results but decreases object detection mAP from 95.3\% to 92.4\% on our synthetic test set. Therefore, our mixture process integrates both cross-resolution attention and addition operations.

\begin{table}[t]
\figcaption{Qualitative boundary visualization results. Semantic-rich features with the lowest resolution lead to blurred boundaries (Fig.~\ref{bun1}), while geometric-rich features with the highest resolution improve clarity (Fig.~\ref{bun2}). The cross-resolution mixture yields the best results (Fig.~\ref{bun5}), compared with using either element-wise addition (Fig.~\ref{bun3}) or MHSA alone (Fig.~\ref{bun4}). Zoom in for the best view.}
\hspace{-0.5cm}
\centering
\begin{subfigure}[t]{0.18\linewidth}
{
\includegraphics[width=1\linewidth,height=1\linewidth]{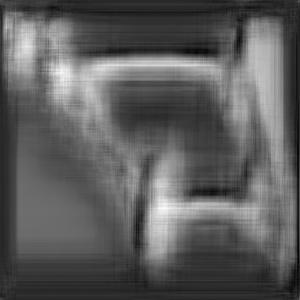}\vspace{-0.1cm}\caption{\label{bun1}}}
\end{subfigure}
\begin{subfigure}[t]{0.18\linewidth}
\includegraphics[width=1\linewidth,height=1\linewidth]{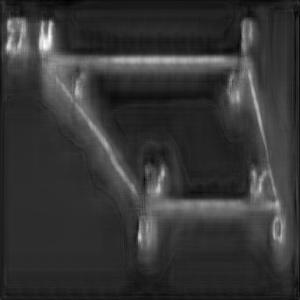}\vspace{-0.1cm}\caption{\label{bun2}}
\end{subfigure}
\begin{subfigure}[t]{0.18\linewidth}
\includegraphics[width=1\linewidth,height=1\linewidth]{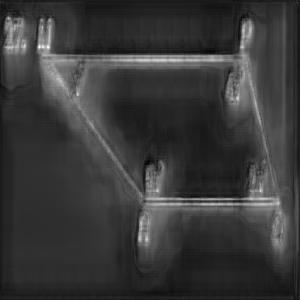}\vspace{-0.1cm}\caption{\label{bun3}}
\end{subfigure}
\begin{subfigure}[t]{0.18\linewidth}
\includegraphics[width=1\linewidth,height=1\linewidth]{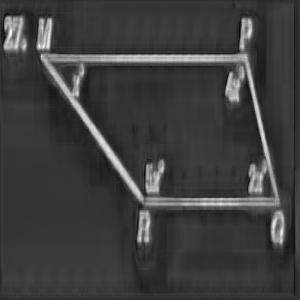}\vspace{-0.1cm}\caption{\label{bun4}}
\end{subfigure}
\begin{subfigure}[t]{0.18\linewidth}
\includegraphics[width=1\linewidth,height=1\linewidth]{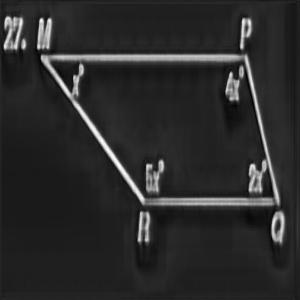}\vspace{-0.1cm}\caption{\label{bun5}}
\end{subfigure}
\vspace{-0.9cm}
\end{table}
\noindent \textbf{Key Factors in Connectors.}
Our connector bridges the soft visual prompts $\widehat{F}_{\text{geo}}$ with the CLIP visual tokens $F_{\text{CLIP}}$ using either channel-wise or sequence-wise fusion methods. We examine two key factors: the inclusion of all visual cues and the use of soft routing. Additionally, for sequence fusion, we explore varying feature resolution sizes. All ablations are conducted on the GeoQA test set. The summary is presented in Fig.~\ref{roadmap}, with detailed top-1 accuracy listed in Fig.~\ref{his_keys}.  Specifically, for smaller resolutions, we resize the pyramid features from \ourgip\ to lengths of 15\%, 20\%, 25\%, and 40\% of the length of $F_{\text{CLIP}}$, respectively, and then sequentially append them to $F_{\text{CLIP}}$. 

Next, we examine the impact of the number of projection experts. The default channel concatenation setup utilizes a single expert with a \texttt{mlp2x\_gelu}. In the multi-expert ablation, where two sequential \texttt{mlp2x\_gelu} are applied, the top-1 accuracy drops from 66.98\% to 64.32\% (-2.66\%), as shown in Fig.\ref{his_keys}. 
Fig.~\ref{his_keys} also shows that the multi-expert setup improves sequence-wise performance over shared parameters (a single expert), increasing accuracy from 64.32\% to 66.58\% (+2.26\%). We hypothesize that the improvement in sequence-wise fusion stems from its added flexibility in handling heterogeneous inputs, while for channel-wise fusion, multi-expert setups may introduce unnecessary complexity and redundancy.


\noindent \textbf{Feature router types and impact of individual feature maps in \ourgip.}
We examine three types of routers: constant, sparse, and the default soft router $R$. The constant router assigns equal weights $w^i = 0.25$ to each $F_{\text{geo}}^i$, while the sparse router selects only one feature map of \ourgip\ with $w^i \in \{0, 1\}$. As expected, in the sparse router, $F_{\text{geo}}^{1^*}$ with more geometric information, achieves the highest accuracy. As shown in Table~\ref{rounter}, the soft router outperforms the others, demonstrating its effectiveness for dynamic routing of multiple signals.

\noindent \textbf{Necessity of CLIP encoder.} While \ourgip\ provides rich geometric visual features, the general visual features provided by models such as CLIP are also crucial. We designed a variant that excludes the CLIP visual encoder, relying solely on our soft prompts from the \ourgip\ visual encoder. Accuracy dropped from 66.6\% to 65.7\% for sequence fusion and from 67.0\% to 66.1\% for channel fusion. These results demonstrate that while CLIP may not perceive fine-grained visual details, its general visual features still benefit text-visual alignment in MLLM training, making such models indispensable in multi-modal mathematical reasoning.

\noindent \textbf{Necessity of GLIP encoder.} GLIP is an open-set object detector capable of identifying arbitrary classes by matching visual features with corresponding language embeddings. Unlike traditional object detectors with learnable classification weights, GLIP's multi-modal architecture offers greater generality to novel objects and surpasses previous traditional object detectors. To evaluate alternatives, we replaced GLIP with another open-set object detector, Grounding DINO~\citep{groundingdino}, and fine-tuned it on our math-specific dataset. Experimentally, we found that Grounding DINO struggles to effectively detect small-scale geometric primitives. We hypothesize that this limitation stems from architectural differences. Grounding DINO, a DETR-based detector, relies on last-layer features for cross-attention with query embeddings, whereas GLIP, a Faster-RCNN-based detector, utilizes multi-scale features for better small-object detection. Integrating the fine-tuned Grounding DINO encoder into our pipeline reduced the top-1 accuracy on the GeoQA benchmark from 67.0\% to 66.1\%, highlighting GLIP's advantages for fine-grained visual perception.

\noindent \textbf{Imapct of math-specific fine-tuning for \ourgip.}  We utilized the hierarchical pyramid features from the GLIP visual encoder (fine-tuned on MS COCO training dataset). To ensure a fair comparison, we utilize the same resolution feature maps: the first layer with the largest resolution and the last three layers with smaller resolutions. This resulted in a drop from 67.0\% to 65.3\%, with only a minimal +1.1\% improvement over G-LLaVA. The slight improvement likely stems from integrating high-resolution vision features, which are not sensitive to geometric details, as GLIP fails to detect basic geometric shapes (Fig. \ref{fig:devis}).

\noindent\textbf{The comparison of visual encoders.} 
\label{encoder-sum}
We have conducted and presented a comprehensive analysis of the effects of various visual encoder variants. For clarity, we provide a summarized overview of their impact in Table \ref{tab:visual_encoders}. We design a variant that excludes the CLIP visual encoder, relying solely on our soft prompts from the GeoGLIP visual encoder. This resulted in an accuracy drop from 67.0\% to 66.1\%, though it still outperformed the CLIP encoder alone (64.2\%). We leveraged the hierarchical pyramid features from the GLIP visual encoder (fine-trained on natural image datasets, such as MS COCO). To ensure a fair comparison, we utilized feature maps with the same resolution: the first layer with the largest resolution and the last three layers with smaller resolutions. This resulted in a performance drop from 67.0\% to 65.3\%, as GLIP lacks sensitivity to geometric details and fails to detect basic geometric shapes, as visualized in Fig. \ref{fig:devis}.


\begin{table}[t]
\vspace{-0.3cm}
\tabcaption{Ablation results \wrt top-1 accuracy on GeoQA. Tab.~\ref{rounter} shows results for feature router types; Fig.~\ref{roadmap} highlights key factors for connector designs, with detailed accuracy in Fig.~\ref{his_keys}.}
\begin{subfigure}[t]{0.2\linewidth}{
\fontsize{7}{10}\selectfont  
\setlength{\tabcolsep}{3pt}
\vspace{-3.cm}
{\begin{tabular}{c|cc||c}
\toprule
& \multicolumn{2}{c||}{Seq.-wise} & Cha.-wise\\
\midrule
Constant $R$& \multicolumn{2}{c||}{63.9}& 62.8\\
\midrule
\multirow{4}{*}{Sparse $R$}& $F_{\text{geo}}^{1^*} \rightarrow$&64.2& $\rightarrow$ 64.9\\
&  $F_{\text{geo}}^{3} \rightarrow$&61.1& $\rightarrow$ 61.8\\
&  $F_{\text{geo}}^{4} \rightarrow$&61.9& $\rightarrow$ 62.3\\
&  $F_{\text{geo}}^{5} \rightarrow$&61.9& $\rightarrow$ 61.6\\
\midrule
Soft $R$& \multicolumn{2}{c||}{\bf 66.6}& \bf 67.0\\
\bottomrule
\end{tabular}}}
\caption{\label{rounter}}
\end{subfigure}
\hspace{2cm}
\begin{subfigure}[t]{0.21\linewidth}
\includegraphics[width=1.1\linewidth,height=1.05\linewidth]{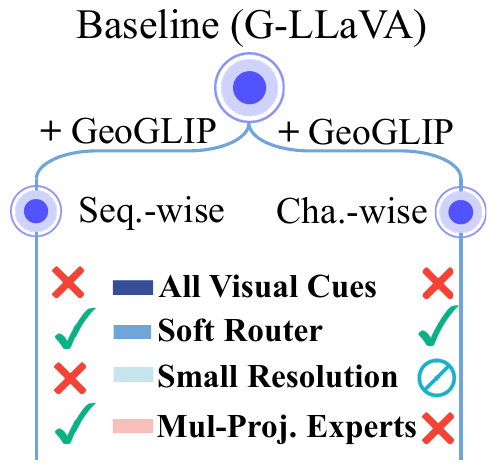}\vspace{0.08cm}\caption{\label{roadmap}}
\end{subfigure}
\hspace{0.5cm}
\begin{subfigure}[t]{0.35\linewidth}
\includegraphics[width=1\linewidth,height=0.68\linewidth]{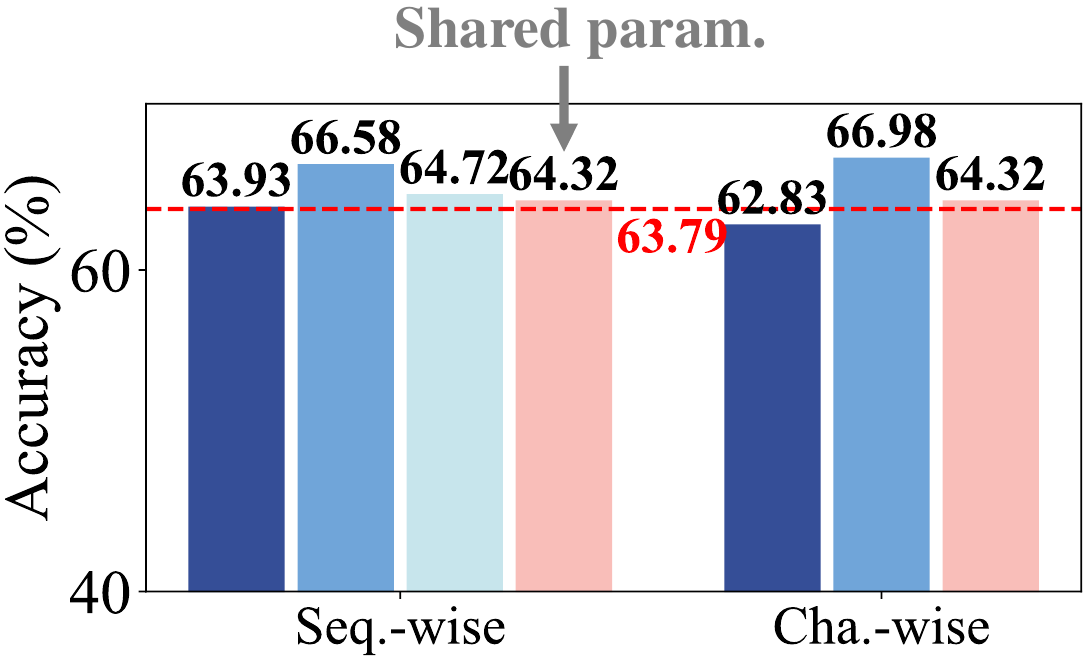}\vspace{0.06cm}\caption{\label{his_keys}}
\end{subfigure}
\vspace{-0.7cm}
\end{table}

\begin{table}[t]
\tabcaption{Comparison of geometric numerical
answer accuracies (\%) on GeoQA \wrt different visual encoder variants (Fig. \ref{tab:visual_encoders}). Fig. \ref{tab:model_comparison} shows top-1 accuracy on testmini set of MathVista \wrt with (SVE-Math) or without our soft visual prompts (SVE-Math(-)).}
\vspace{0.3cm}
\begin{subfigure}[t]{0.48\linewidth}{
\centering
\setlength{\tabcolsep}{30pt}{
    \hspace{0.5cm}
    \setlength{\tabcolsep}{0.05cm}
    \fontsize{7.5}{16}\selectfont
    \begin{tabular}{c|c|c}
        \toprule
        {Type} & {Encoders} & {Accuracy (\%)} \\ 
        \midrule
        Dual encoders & GLIP+CLIP & 65.3 \\
        Dual encoders & \ourgip+CLIP & \bf 67.0 \\ 
        Single encoder & \ourgip & 66.1 \\ 
        Single encoder & CLIP & 64.2 \\ 
        \bottomrule
    \end{tabular}}}
    \caption{\label{tab:visual_encoders}}
    \end{subfigure}
    \hspace{0.5cm}
    \begin{subfigure}[t]{0.45\linewidth}{
    \centering
    \fontsize{7}{10}\selectfont  
    \setlength{\tabcolsep}{3pt}{
    \begin{tabular}{l|c|c}
    \toprule
    {Model} & {Base LLM} & {Acc (All)} \\
    \midrule
    G-LLaVA              & LLaMA2-7B       & 25.1 \\
    {SVE-Math}    & LLaMA2-7B       & 37.4 \\
    \midrule
    SVE-Math-Deepseek(-)          & DeepSeek-math-7B     & 42.3 \\
    {SVE-Math-Deepseek}    & DeepSeek-math-7B     & 48.7 \\
    \midrule
    SVE-Math-Qwen2.5(-)          & Qwen2.5-math-7B      & 44.0 \\
    {SVE-Math-Qwen2.5}    & Qwen2.5-math-7B      & \bf 50.4 \\
    \bottomrule
    \end{tabular}}}
    \caption{\label{tab:model_comparison}}
    \end{subfigure}
    \vspace{-0.8cm}
    \end{table}

\section{Conclusion}

In this paper, we addressed the limitations of current mathematical MLLMs by identifying a key bottleneck: their inability to accurately perceive geometric primitives, essential for mathematical reasoning with visual elements. We proposed \ourMethod, a vision-centric framework that enhances mathematical visual reasoning by integrating a geometry-aware visual encoder trained through multi-task objectives such as shape, junction, and boundary detection. Our approach avoids the labor-intensive process of creating large-scale mathematical visual instruction datasets, offering a more efficient and practical solution. By introducing a feature router that dynamically adjusts the contribution of visual cues, we generate soft prompts that guide the language model to improved reasoning without redundant or irrelevant visual data. Extensive experiments on three public benchmarks demonstrate the effectiveness of \ourMethod, which outperforms similarly sized 7B-parameter models and achieves results comparable to advanced 13B-parameter MLLMs, despite training on smaller datasets. We believe our work highlights the importance of fine-grained visual grounding and adaptive visual cueing mechanisms, providing a foundation for more efficient and interpretable mathematical reasoning in future multimodal models.


\bibliography{SV-Math}

\begin{thebibliography}{75}
\providecommand{\natexlab}[1]{#1}
\providecommand{\url}[1]{\texttt{#1}}
\expandafter\ifx\csname urlstyle\endcsname\relax
  \providecommand{\doi}[1]{doi: #1}\else
  \providecommand{\doi}{doi: \begingroup \urlstyle{rm}\Url}\fi

\bibitem[Academy(2023)]{qwen2023math}
Alibaba~DAMO Academy.
\newblock Qwen: General-purpose large language models with mathematical reasoning abilities, 2023.
\newblock URL \url{https://huggingface.co/Qwen/Qwen-2.5-Math-7B}.

\bibitem[Azerbayev et~al.(2023)Azerbayev, Schoelkopf, Paster, Santos, McAleer, Jiang, Deng, Biderman, and Welleck]{azerbayev2023llemma}
Zhangir Azerbayev, Hailey Schoelkopf, Keiran Paster, Marco~Dos Santos, Stephen McAleer, Albert~Q Jiang, Jia Deng, Stella Biderman, and Sean Welleck.
\newblock Llemma: An open language model for mathematics.
\newblock \emph{arXiv preprint arXiv:2310.10631}, 2023.

\bibitem[Bai et~al.(2023)Bai, Bai, Yang, Wang, Tan, Wang, Lin, Zhou, and Zhou]{bai2023qwen}
Jinze Bai, Shuai Bai, Shusheng Yang, Shijie Wang, Sinan Tan, Peng Wang, Junyang Lin, Chang Zhou, and Jingren Zhou.
\newblock Qwen-vl: A versatile vision-language model for understanding, localization, text reading, and beyond.
\newblock \emph{arXiv preprint arXiv:2308.12966}, 2023.

\bibitem[Cai et~al.(2024{\natexlab{a}})Cai, Bao, Guo, Zhang, Song, and Zheng]{cai2024geogpt4v}
Shihao Cai, Keqin Bao, Hangyu Guo, Jizhi Zhang, Jun Song, and Bo~Zheng.
\newblock Geogpt4v: Towards geometric multi-modal large language models with geometric image generation.
\newblock \emph{arXiv preprint arXiv:2406.11503}, 2024{\natexlab{a}}.

\bibitem[Cai et~al.(2024{\natexlab{b}})Cai, Cao, Chen, Chen, Chen, Chen, Chen, Chen, Chen, Chu, et~al.]{cai2024internlm2}
Zheng Cai, Maosong Cao, Haojiong Chen, Kai Chen, Keyu Chen, Xin Chen, Xun Chen, Zehui Chen, Zhi Chen, Pei Chu, et~al.
\newblock Internlm2 technical report.
\newblock \emph{arXiv preprint arXiv:2403.17297}, 2024{\natexlab{b}}.

\bibitem[Canny(1986)]{canny1986computational}
John Canny.
\newblock A computational approach to edge detection.
\newblock \emph{IEEE Transactions on pattern analysis and machine intelligence}, 8\penalty0 (6):\penalty0 679--698, 1986.

\bibitem[Cao \& Xiao(2022)Cao and Xiao]{cao2022augmented}
Jie Cao and Jing Xiao.
\newblock An augmented benchmark dataset for geometric question answering through dual parallel text encoding.
\newblock In \emph{Proceedings of the 29th International Conference on Computational Linguistics}, pp.\  1511--1520, 2022.

\bibitem[Chen et~al.(2021{\natexlab{a}})Chen, Tang, Qin, Liang, Liu, Xing, and Lin]{chen-etal-2021-geoqa}
Jiaqi Chen, Jianheng Tang, Jinghui Qin, Xiaodan Liang, Lingbo Liu, Eric Xing, and Liang Lin.
\newblock {G}eo{QA}: A geometric question answering benchmark towards multimodal numerical reasoning.
\newblock In Chengqing Zong, Fei Xia, Wenjie Li, and Roberto Navigli (eds.), \emph{Findings of the Association for Computational Linguistics: ACL-IJCNLP 2021}, pp.\  513--523, Online, August 2021{\natexlab{a}}. Association for Computational Linguistics.
\newblock \doi{10.18653/v1/2021.findings-acl.46}.
\newblock URL \url{https://aclanthology.org/2021.findings-acl.46}.

\bibitem[Chen et~al.(2021{\natexlab{b}})Chen, Tang, Qin, Liang, Liu, Xing, and Lin]{chen2021geoqa}
Jiaqi Chen, Jianheng Tang, Jinghui Qin, Xiaodan Liang, Lingbo Liu, Eric~P Xing, and Liang Lin.
\newblock Geoqa: A geometric question answering benchmark towards multimodal numerical reasoning.
\newblock \emph{arXiv preprint arXiv:2105.14517}, 2021{\natexlab{b}}.

\bibitem[Chen et~al.(2022{\natexlab{a}})Chen, Li, Qin, Lu, Lin, Chen, and Liang]{Chen2022UniGeoUG}
Jiaqi Chen, Tong Li, Jinghui Qin, Pan Lu, Liang Lin, Chongyu Chen, and Xiaodan Liang.
\newblock Unigeo: Unifying geometry logical reasoning via reformulating mathematical expression.
\newblock \emph{ArXiv}, abs/2212.02746, 2022{\natexlab{a}}.

\bibitem[Chen et~al.(2022{\natexlab{b}})Chen, Li, Qin, Lu, Lin, Chen, and Liang]{chen2022unigeo}
Jiaqi Chen, Tong Li, Jinghui Qin, Pan Lu, Liang Lin, Chongyu Chen, and Xiaodan Liang.
\newblock {UniGeo}: Unifying geometry logical reasoning via reformulating mathematical expression.
\newblock In \emph{Proceedings of the 2022 Conference on Empirical Methods in Natural Language Processing}, pp.\  3313--3323, 2022{\natexlab{b}}.

\bibitem[Chen et~al.(2023{\natexlab{a}})Chen, Li, Zhang, Xiong, and Elhoseiny]{chen2023minigpt}
Jun Chen, Deyao Zhu1 Xiaoqian Shen1~Xiang Li, Zechun Liu2~Pengchuan Zhang, Raghuraman Krishnamoorthi2 Vikas Chandra2~Yunyang Xiong, and Mohamed Elhoseiny.
\newblock Minigpt-v2: Large language model as a unified interface for vision-language multi-task learning.
\newblock \emph{arXiv preprint arXiv:2310.09478}, 2023{\natexlab{a}}.

\bibitem[Chen et~al.(2023{\natexlab{b}})Chen, Li, wen Dong, Zhang, He, Wang, Zhao, and Lin]{Chen2023ShareGPT4VIL}
Lin Chen, Jinsong Li, Xiao wen Dong, Pan Zhang, Conghui He, Jiaqi Wang, Feng Zhao, and Dahua Lin.
\newblock Sharegpt4v: Improving large multi-modal models with better captions.
\newblock \emph{ArXiv}, abs/2311.12793, 2023{\natexlab{b}}.
\newblock URL \url{https://api.semanticscholar.org/CorpusID:265308687}.

\bibitem[Chiang et~al.(2023)Chiang, Li, Lin, Sheng, Wu, Zhang, Zheng, Zhuang, Zhuang, Gonzalez, Stoica, and Xing]{vicuna2023}
Wei-Lin Chiang, Zhuohan Li, Zi~Lin, Ying Sheng, Zhanghao Wu, Hao Zhang, Lianmin Zheng, Siyuan Zhuang, Yonghao Zhuang, Joseph~E. Gonzalez, Ion Stoica, and Eric~P. Xing.
\newblock Vicuna: An open-source chatbot impressing gpt-4 with 90\%* chatgpt quality.
\newblock \url{https://lmsys.org/blog/2023-03-30-vicuna/}, March 2023.

\bibitem[Dollar et~al.(2006)Dollar, Tu, and Belongie]{dollar2006supervised}
Piotr Dollar, Zhuowen Tu, and Serge Belongie.
\newblock Supervised learning of edges and object boundaries.
\newblock In \emph{2006 IEEE computer society conference on computer vision and pattern recognition (CVPR'06)}, volume~2, pp.\  1964--1971. IEEE, 2006.

\bibitem[Dong et~al.(2024)Dong, Zhang, Zang, Cao, Wang, Ouyang, Wei, Zhang, Duan, Cao, et~al.]{dong2024internlm}
Xiaoyi Dong, Pan Zhang, Yuhang Zang, Yuhang Cao, Bin Wang, Linke Ouyang, Xilin Wei, Songyang Zhang, Haodong Duan, Maosong Cao, et~al.
\newblock Internlm-xcomposer2: Mastering free-form text-image composition and comprehension in vision-language large model.
\newblock \emph{arXiv preprint arXiv:2401.16420}, 2024.

\bibitem[Duda \& Hart(1972)Duda and Hart]{duda1972use}
Richard~O Duda and Peter~E Hart.
\newblock Use of the hough transformation to detect lines and curves in pictures.
\newblock \emph{Communications of the ACM}, 15\penalty0 (1):\penalty0 11--15, 1972.

\bibitem[Gao et~al.(2023{\natexlab{a}})Gao, Pi, Zhang, Ye, Zhong, Wang, Hong, Han, Xu, Li, et~al.]{gao2023g}
Jiahui Gao, Renjie Pi, Jipeng Zhang, Jiacheng Ye, Wanjun Zhong, Yufei Wang, Lanqing Hong, Jianhua Han, Hang Xu, Zhenguo Li, et~al.
\newblock G-llava: Solving geometric problem with multi-modal large language model.
\newblock \emph{arXiv preprint arXiv:2312.11370}, 2023{\natexlab{a}}.

\bibitem[Gao et~al.(2023{\natexlab{b}})Gao, Han, Zhang, Lin, Geng, Zhou, Zhang, Lu, He, Yue, Li, and Qiao]{gao2023llamaadapterv2}
Peng Gao, Jiaming Han, Renrui Zhang, Ziyi Lin, Shijie Geng, Aojun Zhou, Wei Zhang, Pan Lu, Conghui He, Xiangyu Yue, Hongsheng Li, and Yu~Qiao.
\newblock Llama-adapter v2: Parameter-efficient visual instruction model.
\newblock \emph{arXiv preprint arXiv:2304.15010}, 2023{\natexlab{b}}.

\bibitem[Gao et~al.(2024)Gao, Zhang, Liu, Qiu, Huang, Lin, Zhao, Geng, Lin, Jin, et~al.]{gao2024sphinx}
Peng Gao, Renrui Zhang, Chris Liu, Longtian Qiu, Siyuan Huang, Weifeng Lin, Shitian Zhao, Shijie Geng, Ziyi Lin, Peng Jin, et~al.
\newblock Sphinx-x: Scaling data and parameters for a family of multi-modal large language models.
\newblock \emph{arXiv preprint arXiv:2402.05935}, 2024.

\bibitem[Gemini~Team(2023)]{team2023gemini}
Google Gemini~Team.
\newblock Gemini: a family of highly capable multimodal models.
\newblock \emph{arXiv preprint arXiv:2312.11805}, 2023.

\bibitem[Google(2023)]{google2023bard}
Google.
\newblock Bard, 2023.
\newblock URL \url{https://bard.google.com/}.

\bibitem[Gu et~al.(2022)Gu, Lin, Kuo, and Cui]{gu2022openvocabularyobjectdetectionvision}
Xiuye Gu, Tsung-Yi Lin, Weicheng Kuo, and Yin Cui.
\newblock Open-vocabulary object detection via vision and language knowledge distillation, 2022.
\newblock URL \url{https://arxiv.org/abs/2104.13921}.

\bibitem[Han et~al.(2023)Han, Zhang, Shao, Gao, Xu, Xiao, Zhang, Liu, Wen, Guo, et~al.]{han2023imagebind}
Jiaming Han, Renrui Zhang, Wenqi Shao, Peng Gao, Peng Xu, Han Xiao, Kaipeng Zhang, Chris Liu, Song Wen, Ziyu Guo, et~al.
\newblock Imagebind-llm: Multi-modality instruction tuning.
\newblock \emph{arXiv preprint arXiv:2309.03905}, 2023.

\bibitem[Hu et~al.(2024)Hu, Stretcu, Lu, Viswanathan, Hata, Luo, Krishna, and Fuxman]{hu2024visual}
Yushi Hu, Otilia Stretcu, Chun-Ta Lu, Krishnamurthy Viswanathan, Kenji Hata, Enming Luo, Ranjay Krishna, and Ariel Fuxman.
\newblock Visual program distillation: Distilling tools and programmatic reasoning into vision-language models.
\newblock In \emph{Proceedings of the IEEE/CVF Conference on Computer Vision and Pattern Recognition}, pp.\  9590--9601, 2024.

\bibitem[Huang et~al.(2018)Huang, Wang, Zhou, Ding, Gao, and Ma]{huang2018learning}
Kun Huang, Yifan Wang, Zihan Zhou, Tianjiao Ding, Shenghua Gao, and Yi~Ma.
\newblock Learning to parse wireframes in images of man-made environments.
\newblock In \emph{Proceedings of the IEEE Conference on Computer Vision and Pattern Recognition}, pp.\  626--635, 2018.

\bibitem[Jiang et~al.(2024)Jiang, Sablayrolles, Roux, Mensch, Savary, Bamford, Chaplot, de~Las~Casas, Hanna, Bressand, Lengyel, Bour, Lample, Lavaud, Saulnier, Lachaux, Stock, Subramanian, Yang, Antoniak, Scao, Gervet, Lavril, Wang, Lacroix, and Sayed]{albert24mixtral}
Albert~Q. Jiang, Alexandre Sablayrolles, Antoine Roux, Arthur Mensch, Blanche Savary, Chris Bamford, Devendra~Singh Chaplot, Diego de~Las~Casas, Emma~Bou Hanna, Florian Bressand, Gianna Lengyel, Guillaume Bour, Guillaume Lample, L{\'{e}}lio~Renard Lavaud, Lucile Saulnier, Marie{-}Anne Lachaux, Pierre Stock, Sandeep Subramanian, Sophia Yang, Szymon Antoniak, Teven~Le Scao, Th{\'{e}}ophile Gervet, Thibaut Lavril, Thomas Wang, Timoth{\'{e}}e Lacroix, and William~El Sayed.
\newblock Mixtral of experts.
\newblock \emph{Arxiv 2401.04088}, 2024.

\bibitem[Kahou et~al.(2018)Kahou, Michalski, Atkinson, Kadar, Trischler, and Bengio]{figuredataset}
Samira~Ebrahimi Kahou, Vincent Michalski, Adam Atkinson, Akos Kadar, Adam Trischler, and Yoshua Bengio.
\newblock Figureqa: An annotated figure dataset for visual reasoning, 2018.
\newblock URL \url{https://arxiv.org/abs/1710.07300}.

\bibitem[Kazemi et~al.(2023)Kazemi, Alvari, Anand, Wu, Chen, and Soricut]{kazemi2023geomverse}
Mehran Kazemi, Hamidreza Alvari, Ankit Anand, Jialin Wu, Xi~Chen, and Radu Soricut.
\newblock Geomverse: A systematic evaluation of large models for geometric reasoning.
\newblock \emph{arXiv preprint arXiv:2312.12241}, 2023.

\bibitem[Li et~al.(2024)Li, Zhang, Zhang, Guo, Zhang, Li, Zhang, Liu, and Li]{li2024llavanext-strong}
Bo~Li, Kaichen Zhang, Hao Zhang, Dong Guo, Renrui Zhang, Feng Li, Yuanhan Zhang, Ziwei Liu, and Chunyuan Li.
\newblock Llava-next: Stronger llms supercharge multimodal capabilities in the wild, May 2024.
\newblock URL \url{https://llava-vl.github.io/blog/2024-05-10-llava-next-stronger-llms/}.

\bibitem[Li et~al.(2022{\natexlab{a}})Li, Li, Xiong, and Hoi]{li2022blip}
Junnan Li, Dongxu Li, Caiming Xiong, and Steven Hoi.
\newblock Blip: Bootstrapping language-image pre-training for unified vision-language understanding and generation.
\newblock In \emph{International conference on machine learning}, pp.\  12888--12900. PMLR, 2022{\natexlab{a}}.

\bibitem[Li et~al.(2022{\natexlab{b}})Li, Zhang, Zhang, Yang, Li, Zhong, Wang, Yuan, Zhang, Hwang, et~al.]{li2022grounded}
Liunian~Harold Li, Pengchuan Zhang, Haotian Zhang, Jianwei Yang, Chunyuan Li, Yiwu Zhong, Lijuan Wang, Lu~Yuan, Lei Zhang, Jenq-Neng Hwang, et~al.
\newblock Grounded language-image pre-training.
\newblock In \emph{Proceedings of the IEEE/CVF Conference on Computer Vision and Pattern Recognition}, pp.\  10965--10975, 2022{\natexlab{b}}.

\bibitem[Li et~al.(2023)Li, Liu, Wu, Mu, Yang, Gao, Li, and Lee]{li2023gligen}
Yuheng Li, Haotian Liu, Qingyang Wu, Fangzhou Mu, Jianwei Yang, Jianfeng Gao, Chunyuan Li, and Yong~Jae Lee.
\newblock Gligen: Open-set grounded text-to-image generation.
\newblock In \emph{Proceedings of the IEEE/CVF Conference on Computer Vision and Pattern Recognition}, pp.\  22511--22521, 2023.

\bibitem[Liang et~al.(2023)Liang, Yang, Zhang, and Zhang]{liang_unimath}
Zhenwen Liang, Tianyu Yang, Jipeng Zhang, and Xiangliang Zhang.
\newblock Unimath: A foundational and multimodal mathematical reasoner.
\newblock In \emph{EMNLP}, 2023.

\bibitem[Lin et~al.(2023)Lin, Liu, Zhang, Gao, Qiu, Xiao, Qiu, Lin, Shao, Chen, et~al.]{lin2023sphinx}
Ziyi Lin, Chris Liu, Renrui Zhang, Peng Gao, Longtian Qiu, Han Xiao, Han Qiu, Chen Lin, Wenqi Shao, Keqin Chen, et~al.
\newblock Sphinx: The joint mixing of weights, tasks, and visual embeddings for multi-modal large language models.
\newblock \emph{arXiv preprint arXiv:2311.07575}, 2023.

\bibitem[Liu et~al.(2023{\natexlab{a}})Liu, Li, Li, and Lee]{liu2023improvedllava}
Haotian Liu, Chunyuan Li, Yuheng Li, and Yong~Jae Lee.
\newblock Improved baselines with visual instruction tuning, 2023{\natexlab{a}}.

\bibitem[Liu et~al.(2023{\natexlab{b}})Liu, Li, Wu, and Lee]{liu2023llava}
Haotian Liu, Chunyuan Li, Qingyang Wu, and Yong~Jae Lee.
\newblock Visual instruction tuning.
\newblock In \emph{NeurIPS}, 2023{\natexlab{b}}.

\bibitem[Liu et~al.(2024{\natexlab{a}})Liu, Li, Li, Li, Zhang, Shen, and Lee]{liu2024llavanext}
Haotian Liu, Chunyuan Li, Yuheng Li, Bo~Li, Yuanhan Zhang, Sheng Shen, and Yong~Jae Lee.
\newblock Llava-next: Improved reasoning, ocr, and world knowledge, January 2024{\natexlab{a}}.
\newblock URL \url{https://llava-vl.github.io/blog/2024-01-30-llava-next/}.

\bibitem[Liu et~al.(2024{\natexlab{b}})Liu, Li, Wu, and Lee]{liu2024visual}
Haotian Liu, Chunyuan Li, Qingyang Wu, and Yong~Jae Lee.
\newblock Visual instruction tuning.
\newblock \emph{Advances in neural information processing systems}, 36, 2024{\natexlab{b}}.

\bibitem[Liu et~al.(2024{\natexlab{c}})Liu, Zeng, Ren, Li, Zhang, Yang, Jiang, Li, Yang, Su, Zhu, and Zhang]{groundingdino}
Shilong Liu, Zhaoyang Zeng, Tianhe Ren, Feng Li, Hao Zhang, Jie Yang, Qing Jiang, Chunyuan Li, Jianwei Yang, Hang Su, Jun Zhu, and Lei Zhang.
\newblock Grounding dino: Marrying dino with grounded pre-training for open-set object detection, 2024{\natexlab{c}}.
\newblock URL \url{https://arxiv.org/abs/2303.05499}.

\bibitem[Liu et~al.(2024{\natexlab{d}})Liu, Zeng, Ren, Li, Zhang, Yang, Jiang, Li, Yang, Su, Zhu, and Zhang]{liu2024groundingdinomarryingdino}
Shilong Liu, Zhaoyang Zeng, Tianhe Ren, Feng Li, Hao Zhang, Jie Yang, Qing Jiang, Chunyuan Li, Jianwei Yang, Hang Su, Jun Zhu, and Lei Zhang.
\newblock Grounding dino: Marrying dino with grounded pre-training for open-set object detection, 2024{\natexlab{d}}.
\newblock URL \url{https://arxiv.org/abs/2303.05499}.

\bibitem[Lu et~al.(2023)Lu, Bansal, Xia, Liu, yue Li, Hajishirzi, Cheng, Chang, Galley, and Gao]{Lu2023MathVistaEM}
Pan Lu, Hritik Bansal, Tony Xia, Jiacheng Liu, Chun yue Li, Hannaneh Hajishirzi, Hao Cheng, Kai-Wei Chang, Michel Galley, and Jianfeng Gao.
\newblock Mathvista: Evaluating math reasoning in visual contexts with gpt-4v, bard, and other large multimodal models.
\newblock \emph{ArXiv}, abs/2310.02255, 2023.

\bibitem[Luo et~al.(2023)Luo, Sun, Xu, Zhao, Lou, Tao, Geng, Lin, Chen, and Zhang]{luo2023wizardmath}
Haipeng Luo, Qingfeng Sun, Can Xu, Pu~Zhao, Jianguang Lou, Chongyang Tao, Xiubo Geng, Qingwei Lin, Shifeng Chen, and Dongmei Zhang.
\newblock Wizardmath: Empowering mathematical reasoning for large language models via reinforced evol-instruct.
\newblock \emph{arXiv preprint arXiv:2308.09583}, 2023.

\bibitem[Maire et~al.(2008)Maire, Arbelaez, Fowlkes, and Malik]{maire2008using}
Michael Maire, Pablo Arbelaez, Charless Fowlkes, and Jitendra Malik.
\newblock Using contours to detect and localize junctions in natural images.
\newblock In \emph{2008 IEEE Conference on Computer Vision and Pattern Recognition}, pp.\  1--8. IEEE, 2008.

\bibitem[Ning et~al.(2023)Ning, Wang, Huang, and Huang]{sca_gps}
Maizhen Ning, Qiu-Feng Wang, Kaizhu Huang, and Xiaowei Huang.
\newblock A symbolic characters aware model for solving geometry problems.
\newblock In \emph{Proceedings of the 31st ACM International Conference on Multimedia}, MM '23, pp.\  7767–7775, New York, NY, USA, 2023. Association for Computing Machinery.
\newblock ISBN 9798400701085.
\newblock \doi{10.1145/3581783.3612570}.
\newblock URL \url{https://doi.org/10.1145/3581783.3612570}.

\bibitem[OpenAI(2023{\natexlab{a}})]{OpenAI2023ChatGPT}
OpenAI.
\newblock Chatgpt.
\newblock \url{https://chat.openai.com}, 2023{\natexlab{a}}.

\bibitem[OpenAI(2023{\natexlab{b}})]{OpenAI2023GPT4TR}
OpenAI.
\newblock Gpt-4 technical report.
\newblock \emph{ArXiv}, abs/2303.08774, 2023{\natexlab{b}}.

\bibitem[OpenAI(2023{\natexlab{c}})]{openai2023gpt4v}
OpenAI.
\newblock {GPT-4V(ision)} system card, 2023{\natexlab{c}}.
\newblock URL \url{https://openai.com/research/gpt-4v-system-card}.

\bibitem[Ouyang et~al.(2022)Ouyang, Wu, Jiang, Almeida, Wainwright, Mishkin, Zhang, Agarwal, Slama, Gray, Schulman, Hilton, Kelton, Miller, Simens, Askell, Welinder, Christiano, Leike, and Lowe]{ouyang2022training}
Long Ouyang, Jeffrey Wu, Xu~Jiang, Diogo Almeida, Carroll Wainwright, Pamela Mishkin, Chong Zhang, Sandhini Agarwal, Katarina Slama, Alex Gray, John Schulman, Jacob Hilton, Fraser Kelton, Luke Miller, Maddie Simens, Amanda Askell, Peter Welinder, Paul Christiano, Jan Leike, and Ryan Lowe.
\newblock Training language models to follow instructions with human feedback.
\newblock In \emph{Advances in Neural Information Processing Systems}, 2022.

\bibitem[Parida et~al.(1998)Parida, Geiger, and Hummel]{parida1998junctions}
Laxmi Parida, Davi Geiger, and Robert Hummel.
\newblock Junctions: Detection, classification, and reconstruction.
\newblock \emph{IEEE Transactions on Pattern Analysis and Machine Intelligence}, 20\penalty0 (7):\penalty0 687--698, 1998.

\bibitem[Peng et~al.(2023)Peng, Wang, Dong, Hao, Huang, Ma, and Wei]{peng2023kosmos}
Zhiliang Peng, Wenhui Wang, Li~Dong, Yaru Hao, Shaohan Huang, Shuming Ma, and Furu Wei.
\newblock Kosmos-2: Grounding multimodal large language models to the world.
\newblock \emph{arXiv preprint arXiv:2306.14824}, 2023.

\bibitem[Puigcerver et~al.(2024)Puigcerver, Riquelme, Mustafa, and Houlsby]{softmixturesexperts}
Joan Puigcerver, Carlos Riquelme, Basil Mustafa, and Neil Houlsby.
\newblock From sparse to soft mixtures of experts, 2024.
\newblock URL \url{https://arxiv.org/abs/2308.00951}.

\bibitem[Radford et~al.(2021)Radford, Kim, Hallacy, Ramesh, Goh, Agarwal, Sastry, Askell, Mishkin, Clark, Krueger, and Sutskever]{Radford2021LearningTV}
Alec Radford, Jong~Wook Kim, Chris Hallacy, Aditya Ramesh, Gabriel Goh, Sandhini Agarwal, Girish Sastry, Amanda Askell, Pamela Mishkin, Jack Clark, Gretchen Krueger, and Ilya Sutskever.
\newblock Learning transferable visual models from natural language supervision.
\newblock In \emph{International Conference on Machine Learning}, 2021.
\newblock URL \url{https://api.semanticscholar.org/CorpusID:231591445}.

\bibitem[Shi et~al.(2024)Shi, Hu, Bin, Liu, Yang, Ng, Bing, and Lee]{shi2024math}
Wenhao Shi, Zhiqiang Hu, Yi~Bin, Junhua Liu, Yang Yang, See-Kiong Ng, Lidong Bing, and Roy Ka-Wei Lee.
\newblock Math-llava: Bootstrapping mathematical reasoning for multimodal large language models.
\newblock \emph{arXiv preprint arXiv:2406.17294}, 2024.

\bibitem[Su et~al.(2023)Su, Lan, Li, Xu, Wang, and Cai]{su2023pandagpt}
Yixuan Su, Tian Lan, Huayang Li, Jialu Xu, Yan Wang, and Deng Cai.
\newblock Pandagpt: One model to instruction-follow them all.
\newblock \emph{arXiv preprint arXiv:2305.16355}, 2023.

\bibitem[Sur{\'\i}s et~al.(2023)Sur{\'\i}s, Menon, and Vondrick]{suris2023vipergpt}
D{\'\i}dac Sur{\'\i}s, Sachit Menon, and Carl Vondrick.
\newblock Vipergpt: Visual inference via python execution for reasoning.
\newblock In \emph{Proceedings of the IEEE/CVF International Conference on Computer Vision}, pp.\  11888--11898, 2023.

\bibitem[Team(2023)]{deepseek2023math}
DeepSeek Team.
\newblock Deepseek: Advanced mathematical reasoning for large language models, 2023.
\newblock URL \url{https://huggingface.co/DeepSeek/DeepSeek-Math-7B-Instruct}.

\bibitem[Team(2024)]{qwen1.5}
Qwen Team.
\newblock Introducing qwen1.5, February 2024.
\newblock URL \url{https://qwenlm.github.io/blog/qwen1.5/}.

\bibitem[Tong et~al.(2024)Tong, Liu, Zhai, Ma, LeCun, and Xie]{tong2024eyes}
Shengbang Tong, Zhuang Liu, Yuexiang Zhai, Yi~Ma, Yann LeCun, and Saining Xie.
\newblock Eyes wide shut? exploring the visual shortcomings of multimodal llms.
\newblock In \emph{Proceedings of the IEEE/CVF Conference on Computer Vision and Pattern Recognition}, pp.\  9568--9578, 2024.

\bibitem[Touvron et~al.(2023{\natexlab{a}})Touvron, Lavril, Izacard, Martinet, Lachaux, Lacroix, Rozi{\`e}re, Goyal, Hambro, Azhar, et~al.]{touvron2023llama}
Hugo Touvron, Thibaut Lavril, Gautier Izacard, Xavier Martinet, Marie-Anne Lachaux, Timoth{\'e}e Lacroix, Baptiste Rozi{\`e}re, Naman Goyal, Eric Hambro, Faisal Azhar, et~al.
\newblock Llama: Open and efficient foundation language models.
\newblock \emph{arXiv preprint arXiv:2302.13971}, 2023{\natexlab{a}}.

\bibitem[Touvron et~al.(2023{\natexlab{b}})Touvron, Martin, Stone, Albert, Almahairi, Babaei, Bashlykov, Batra, Bhargava, Bhosale, et~al.]{touvron2023llama2}
Hugo Touvron, Louis Martin, Kevin Stone, Peter Albert, Amjad Almahairi, Yasmine Babaei, Nikolay Bashlykov, Soumya Batra, Prajjwal Bhargava, Shruti Bhosale, et~al.
\newblock Llama 2: Open foundation and fine-tuned chat models.
\newblock \emph{arXiv preprint arXiv:2307.09288}, 2023{\natexlab{b}}.

\bibitem[Verbin \& Zickler(2021)Verbin and Zickler]{verbin2021field}
Dor Verbin and Todd Zickler.
\newblock Field of junctions: Extracting boundary structure at low snr.
\newblock In \emph{Proceedings of the IEEE/CVF International Conference on Computer Vision}, pp.\  6869--6878, 2021.

\bibitem[Wang et~al.(2024)Wang, Pan, Shi, Lu, Zhan, and Li]{wang2024measuring}
Ke~Wang, Junting Pan, Weikang Shi, Zimu Lu, Mingjie Zhan, and Hongsheng Li.
\newblock Measuring multimodal mathematical reasoning with math-vision dataset.
\newblock \emph{arXiv preprint arXiv:2402.14804}, 2024.

\bibitem[Yao et~al.(2022)Yao, Han, Wen, Liang, Xu, Zhang, Li, Xu, and Xu]{yao2022detclipdictionaryenrichedvisualconceptparalleled}
Lewei Yao, Jianhua Han, Youpeng Wen, Xiaodan Liang, Dan Xu, Wei Zhang, Zhenguo Li, Chunjing Xu, and Hang Xu.
\newblock Detclip: Dictionary-enriched visual-concept paralleled pre-training for open-world detection, 2022.
\newblock URL \url{https://arxiv.org/abs/2209.09407}.

\bibitem[Ye et~al.(2023{\natexlab{a}})Ye, Xu, Xu, Ye, Yan, Zhou, Wang, Hu, Shi, Shi, Jiang, Li, Xu, Chen, Tian, Qian, Zhang, and Huang]{ye2023mplugowl}
Qinghao Ye, Haiyang Xu, Guohai Xu, Jiabo Ye, Ming Yan, Yiyang Zhou, Junyang Wang, Anwen Hu, Pengcheng Shi, Yaya Shi, Chaoya Jiang, Chenliang Li, Yuanhong Xu, Hehong Chen, Junfeng Tian, Qi~Qian, Ji~Zhang, and Fei Huang.
\newblock mplug-owl: Modularization empowers large language models with multimodality, 2023{\natexlab{a}}.

\bibitem[Ye et~al.(2023{\natexlab{b}})Ye, Xu, Ye, Yan, Hu, Liu, Qian, Zhang, Huang, and Zhou]{ye2023mplugowl2}
Qinghao Ye, Haiyang Xu, Jiabo Ye, Ming Yan, Anwen Hu, Haowei Liu, Qi~Qian, Ji~Zhang, Fei Huang, and Jingren Zhou.
\newblock mplug-owl2: Revolutionizing multi-modal large language model with modality collaboration, 2023{\natexlab{b}}.

\bibitem[Ying et~al.(2024)Ying, Zhang, Li, Zhou, Shao, Fei, Ma, Hong, Liu, Wang, et~al.]{ying2024internlm}
Huaiyuan Ying, Shuo Zhang, Linyang Li, Zhejian Zhou, Yunfan Shao, Zhaoye Fei, Yichuan Ma, Jiawei Hong, Kuikun Liu, Ziyi Wang, et~al.
\newblock Internlm-math: Open math large language models toward verifiable reasoning.
\newblock \emph{arXiv preprint arXiv:2402.06332}, 2024.

\bibitem[Yu et~al.(2023)Yu, Jiang, Shi, Yu, Liu, Zhang, Kwok, Li, Weller, and Liu]{yu2023metamath}
Longhui Yu, Weisen Jiang, Han Shi, Jincheng Yu, Zhengying Liu, Yu~Zhang, James~T Kwok, Zhenguo Li, Adrian Weller, and Weiyang Liu.
\newblock Metamath: Bootstrap your own mathematical questions for large language models.
\newblock \emph{arXiv preprint arXiv:2309.12284}, 2023.

\bibitem[Yue et~al.(2023{\natexlab{a}})Yue, Ni, Zhang, Zheng, Liu, Zhang, Stevens, Jiang, Ren, Sun, Wei, Yu, Yuan, Sun, Yin, Zheng, Yang, Liu, Huang, Sun, Su, and Chen]{yue2023mmmu}
Xiang Yue, Yuansheng Ni, Kai Zhang, Tianyu Zheng, Ruoqi Liu, Ge~Zhang, Samuel Stevens, Dongfu Jiang, Weiming Ren, Yuxuan Sun, Cong Wei, Botao Yu, Ruibin Yuan, Renliang Sun, Ming Yin, Boyuan Zheng, Zhenzhu Yang, Yibo Liu, Wenhao Huang, Huan Sun, Yu~Su, and Wenhu Chen.
\newblock Mmmu: A massive multi-discipline multimodal understanding and reasoning benchmark for expert agi.
\newblock \emph{arXiv preprint arXiv:2311.16502}, 2023{\natexlab{a}}.

\bibitem[Yue et~al.(2023{\natexlab{b}})Yue, Qu, Zhang, Fu, Huang, Sun, Su, and Chen]{yue2023mammoth}
Xiang Yue, Xingwei Qu, Ge~Zhang, Yao Fu, Wenhao Huang, Huan Sun, Yu~Su, and Wenhu Chen.
\newblock Mammoth: Building math generalist models through hybrid instruction tuning.
\newblock \emph{arXiv preprint arXiv:2309.05653}, 2023{\natexlab{b}}.

\bibitem[Yue et~al.(2024)Yue, Zheng, Zhang, and Chen]{yue2024mammoth2}
Xiang Yue, Tuney Zheng, Ge~Zhang, and Wenhu Chen.
\newblock Mammoth2: Scaling instructions from the web.
\newblock \emph{arXiv preprint arXiv:2405.03548}, 2024.

\bibitem[Zareian et~al.(2021)Zareian, Rosa, Hu, and Chang]{zareian2021openvocabularyobjectdetectionusing}
Alireza Zareian, Kevin~Dela Rosa, Derek~Hao Hu, and Shih-Fu Chang.
\newblock Open-vocabulary object detection using captions, 2021.
\newblock URL \url{https://arxiv.org/abs/2011.10678}.

\bibitem[Zhang et~al.(2024{\natexlab{a}})Zhang, Jiang, Zhang, Lin, Guo, Qiu, Zhou, Lu, Chang, Gao, et~al.]{zhang2024mathverse}
Renrui Zhang, Dongzhi Jiang, Yichi Zhang, Haokun Lin, Ziyu Guo, Pengshuo Qiu, Aojun Zhou, Pan Lu, Kai-Wei Chang, Peng Gao, et~al.
\newblock Mathverse: Does your multi-modal llm truly see the diagrams in visual math problems?
\newblock \emph{arXiv preprint arXiv:2403.14624}, 2024{\natexlab{a}}.

\bibitem[Zhang et~al.(2024{\natexlab{b}})Zhang, Wei, Jiang, Zhang, Guo, Tong, Liu, Zhou, Wei, Zhang, et~al.]{zhang2024mavis}
Renrui Zhang, Xinyu Wei, Dongzhi Jiang, Yichi Zhang, Ziyu Guo, Chengzhuo Tong, Jiaming Liu, Aojun Zhou, Bin Wei, Shanghang Zhang, et~al.
\newblock Mavis: Mathematical visual instruction tuning.
\newblock \emph{arXiv preprint arXiv:2407.08739}, 2024{\natexlab{b}}.

\bibitem[Zhu et~al.(2023)Zhu, Chen, Shen, Li, and Elhoseiny]{zhu2023minigpt}
Deyao Zhu, Jun Chen, Xiaoqian Shen, Xiang Li, and Mohamed Elhoseiny.
\newblock Minigpt-4: Enhancing vision-language understanding with advanced large language models.
\newblock \emph{arXiv preprint arXiv:2304.10592}, 2023.

\end{thebibliography}
\bibliographystyle{SV-Math}
\clearpage
\appendix
\section*{{\raisebox{-0.2\height}{\includegraphics[scale=0.04]{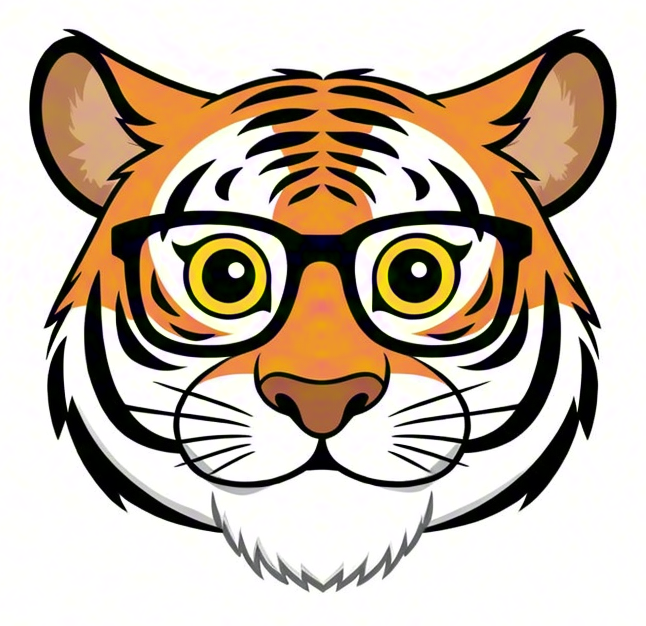}}}  Open Eyes, Then Reason: Fine-grained Visual Mathematical Understanding in MLLMs}
\addcontentsline{toc}{section}{Supplementary Material}
\section{Appendix}
\label{supp}
In this supplementary material, we illustrate the related background for our method (\S~\ref{sec:bg}), provide a detailed description for \ourgip\ (Geometric-Grounded Language-Image Pre-training) pipeline (\S~\ref{detepipe}), explain the process of synthetic data generation, and outline the datasets used for training \ourgip\ (\S~\ref{syndata}), present visualizations of the \ourgip\ detection results (\S~\ref{detevis}),  offer case studies that illustrate the practical application of our feature router mechanism and chain-of-thought (CoT) reasoning results (\S~\ref{case}), demonstrate the training details/efficiency of \ourMethod\ (\S~\ref{cost}) and examine our model's limitations while outlining potential directions for future work (\S~\ref{furwork}).


\subsection{Background}
\label{sec:bg}
\noindent \textbf{Grounded Language-Image Pre-training (GLIP).} 
GLIP \citep{li2022grounded} unifies detection and grounding by reformulating object detection as phrase grounding. It accepts paired image-text inputs, where the text consists of candidate detection categories, such as the 80 COCO object class names joined by `.', \ie, person. bicycle. car. $\cdots$ toothbrush. In GLIP, object classification logits in the box classifier (traditional object detection) are replaced with word-region alignment scores, computed as the dot product between region visual features and phrase language features. GLIP operates as a two-stage detector, composed of: 1) A Swin Transformer as a visual encoder, which extracts features $F_I$ of images $X_I$  and passes $F_I$ to a Region Proposal Network (RPN) to generate region coordinates, and then corresponding region features $O_I$ are cropped from $F_I$; 2) A pre-trained BERT model as the language encoder, to embed the input text  $X_L$ into token embeddings $P_L$; 3) A language-aware deep fusion module Fus$_{IL}$ that fuses $O_I$ and $P_L$ in the last few encoding layers. The final alignment scores $S_\text{ground}$, calculated as:
\[
O_I = \text{RPN}(\text{Swin}(X_I)), \quad P_L = \text{BERT}(X_L), \quad O_I',P_L'=\text{Fus}_{IL}(O_I,P_L) \quad S_\text{ground} = O_I',P_L'^\top. 
\]
\noindent \textbf{Large Language and Vision Assistant (LLaVA).} 
We adopt (Large Language and Vision Assistant) LLaVA’s architecture ~\citep{liu2023llava}  as the basis.
LLaVA  leverages the complementary strengths of pre-trained large language models and visual encoders to perform multi-modal tasks, consisting of a large language model $f_\phi$ (Vicuna~\citep{vicuna2023}), a vision encoder (CLIP, ViT-L/14)~\citep{Radford2021LearningTV}, and a projection layer. The projection layer projects the visual embedding from the vision encoder into the text embedding space.  LLaVA begins by processing an input image $X_I$ through the CLIP visual encoder, which extracts visual features ${F_I} = \text{CLIP}(X_I)$. To bridge the gap between the image features and the language model's word embedding space, LLaVA applies a simple linear projection matrix $\boldsymbol{\Phi}$, converting visual features ${F_I}$ into visual tokens ${H_I}$, which are compatible with the language embedding space:
\[
{H_I} ={\boldsymbol{\Phi}} \cdot {F_I}, \text{ with } {F_I} = \text{CLIP}({X_I})
\]
The visual tokens ${H_I}$ and language instruction tokens $P_L$ are passed into the language model for joint reasoning and language generation as $f_\phi ([{H_I},P_L])$.

\subsection{GeoGLIP}
\label{detepipe}
\begin{figure}[t]
    \centering
    \vspace{-0.8cm}
    \includegraphics[width=0.8\linewidth]{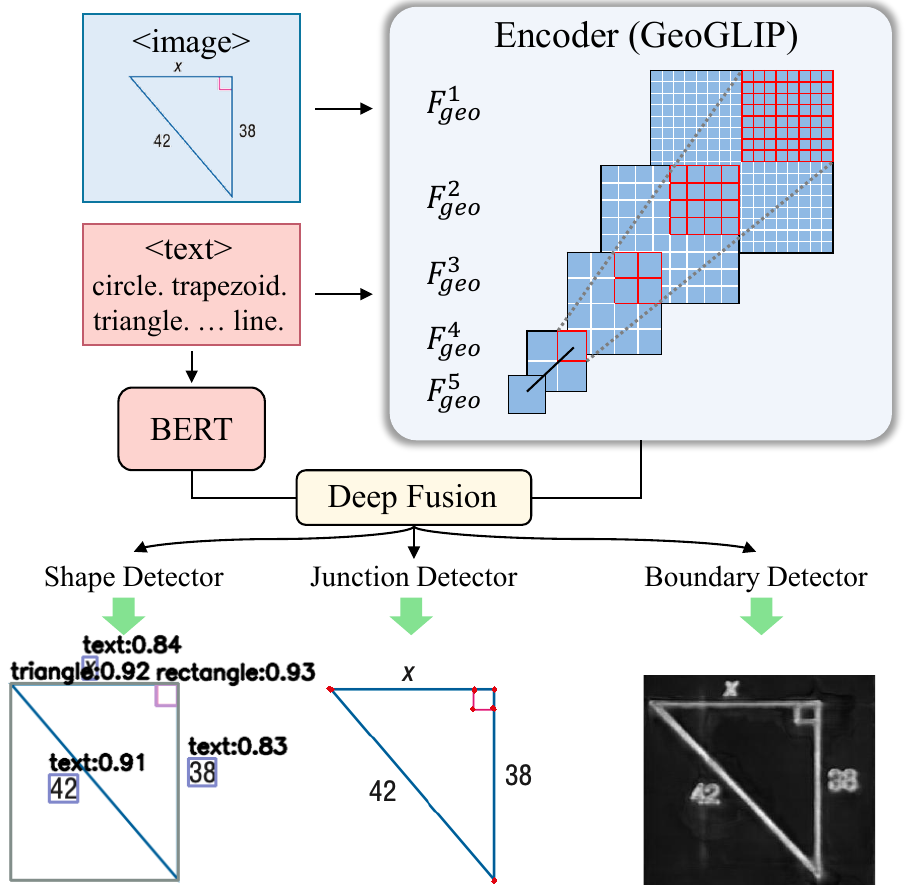}\vspace{-0.2cm}
    \figcaption{\ourgip\ pipeline. A geometric multi-task detector. \ourgip\ simultaneously detects multiple tasks, including basic geometric shapes, junctions, and boundaries, utilizing multi-scale features to capture fine-grained geometric entities. \vspace{-0.3cm}}
    \label{fig:glip}
\end{figure}

The \ourgip\ pipeline is shown in Fig.~\ref{fig:glip}, where the RPN and language-aware deep fusion details are omitted for clarity. The \ourgip\ takes image-text paired as input: an image containing geometric shapes and a text listing the shape classes (\ie, `circle. trapezoid. triangle. $\dots$ line.'). These inputs are processed by the \ourgip\ encoder, which generates feature pyramids at multiple scales ($F_{\text{geo}}^{1}, F_{\text{geo}}^{2}, F_{\text{geo}}^{3}, F_{\text{geo}}^{4}, F_{\text{geo}}^{5}$). Each feature pyramid contains different levels of detail, capturing varying levels of geometric information. These features are then routed to three separate detectors: 1) Shape Detector: identifies and localizes basic geometric shapes by generating bounding boxes for objects within the image; 2) Junction Detector: detects junctions or intersections of geometric entities in the image; 3) Boundary Detector: identifies boundaries of geometric shapes, refining their outlines for more accurate representation. The combination of the feature pyramids  with task-specific detectors allows \ourgip\ to perform fine-grained visual tasks in a mathematical context.  

In Fig.~\ref{fig:devis}, we illustrate detailed designs about junction and boundary detectors:
\begin{itemize}
\vspace{-0.2cm}
\item Junction Detector: The detector processes the feature  $F_{\text{geo}}^{1^*}$ through a decoder, identifying the confidence of junction points within each grid cell and their relative positions.
It also predicts the orientations and confidence levels of intersecting lines within the grid, split into multiple angular bins to cover the 360-degree range.
\item Boundary Detector: It employs two successive perception blocks and upsampling operations to restore the feature map to the original image resolution for boundary decoding.
\vspace{-0.5cm}
\end{itemize}
Both detectors use multi-resolution feature maps from the \ourgip\ encoder, and specific design for  each task is optimized to capture relevant geometric properties, contributing to enhanced mathematical visual reasoning. 
Refer to \S~\ref{MAGLIP} of main paper for more details.
\begin{figure}[t]
    \centering
    
    \begin{subfigure}[b]{0.32\linewidth}
        \centering
        \includegraphics[width=1\linewidth]{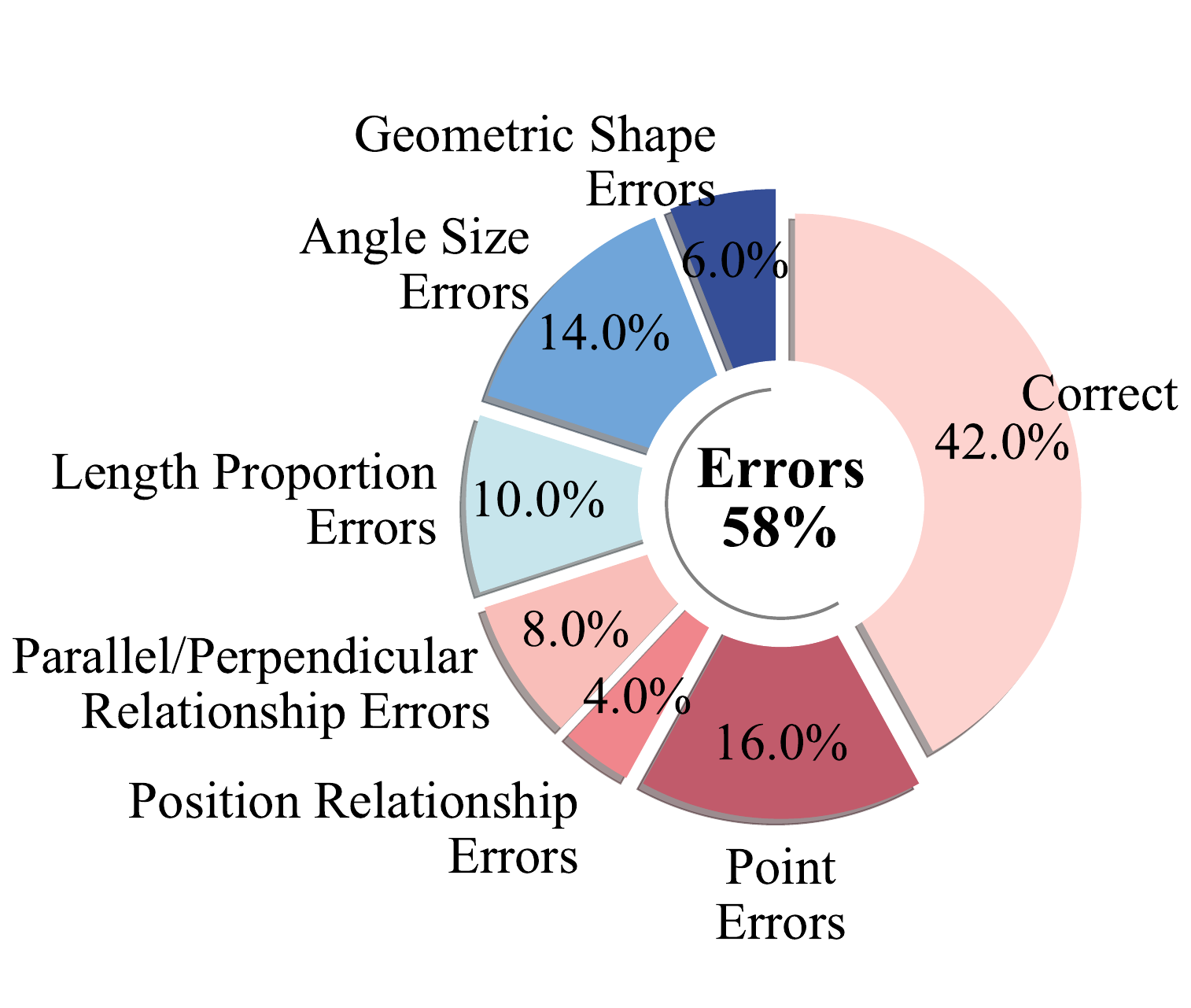}
        \caption{}
        \label{fig:supp_subfig1}
    \end{subfigure}
    \hfill
    \begin{subfigure}[b]{0.32\linewidth}
        \centering
        \includegraphics[width=\linewidth]{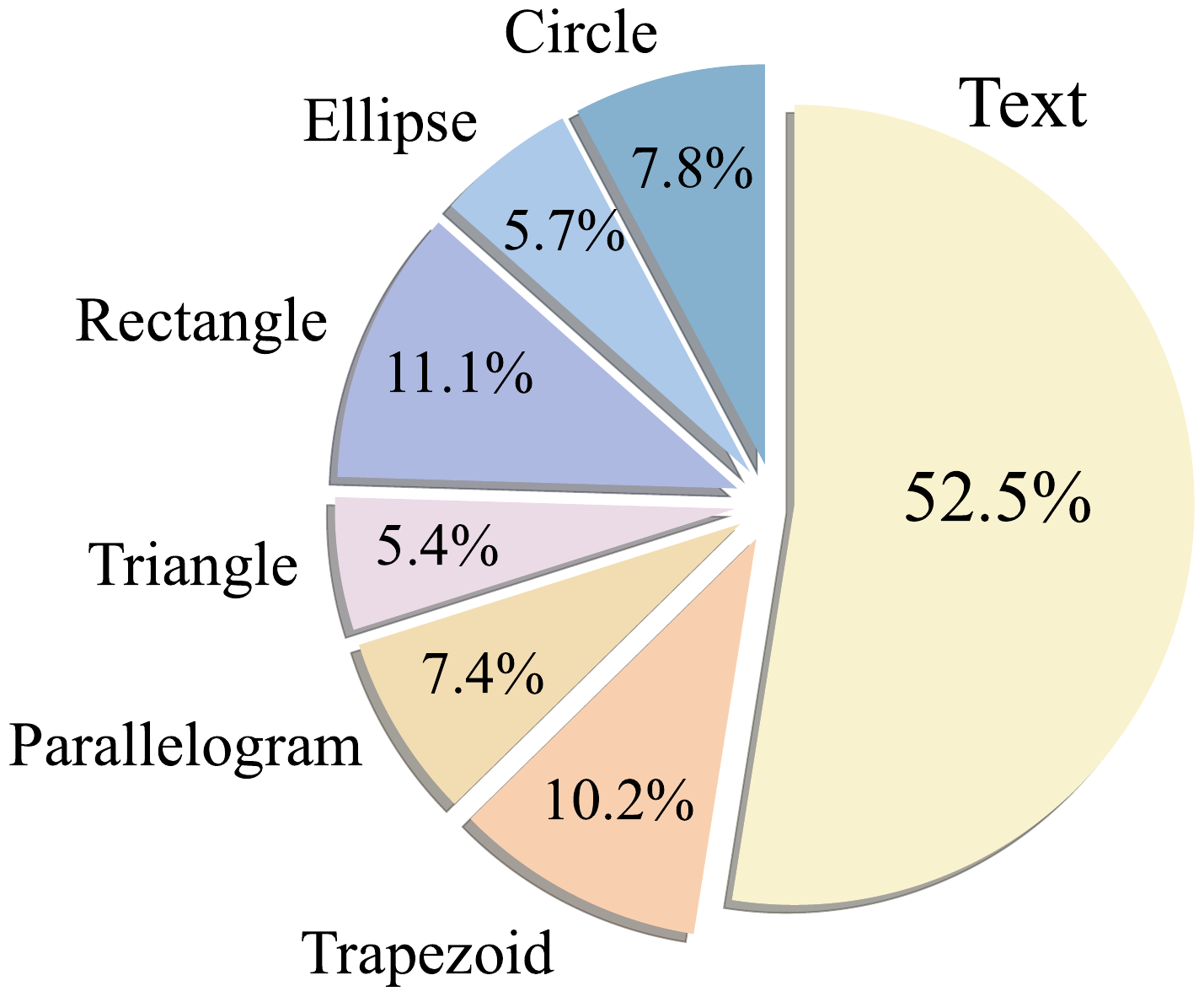}
        \caption{}
        \label{fig:supp_subfig2}
    \end{subfigure}
    \hfill
    \begin{subfigure}[b]{0.32\linewidth}
        \centering
        \includegraphics[width=\linewidth]{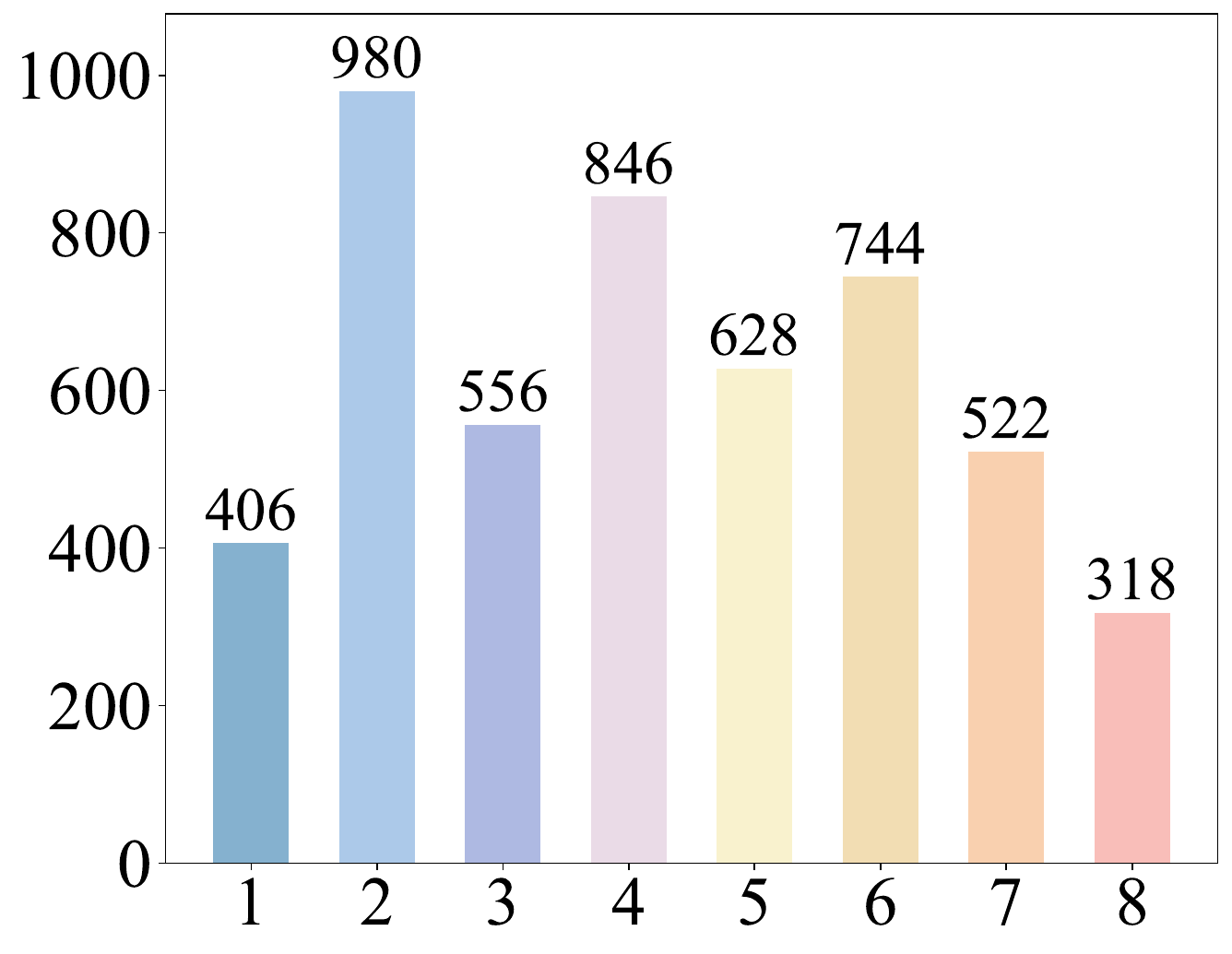}
        \caption{}
        \label{fig:supp_subfig3}
    \end{subfigure}
    
    \caption{Fig.~\ref{fig:supp_subfig1} presents the statistics of top-1 accuracy after manually correcting the visual perception errors shown in Fig.\ref{fig:intro_subfig1} of the main paper, which initially caused incorrect answers to mathematical questions. Specifically, we restated the output of GPT-4o \wrt each type of visual recognition error and calculated the accuracy of its answers. Overall, correcting these visual perception errors led to an approximate 12\% increase in accuracy on the corresponding mathematical questions. Fig.~\ref{fig:supp_subfig2} and  Fig.~\ref{fig:supp_subfig3} present the data statistics for synthetic math-specific datasets, including the distribution of geometric shapes/classes and the number of objects per image. Each geometric object has a 70\% probability of being assigned an alphanumeric text, leading to a higher proportion of the `Text' class. }
    \vspace{-0.2cm}
    \label{fig:supp:data}
\end{figure}
\begin{figure}[t]
    \centering
    \includegraphics[width=1\linewidth]{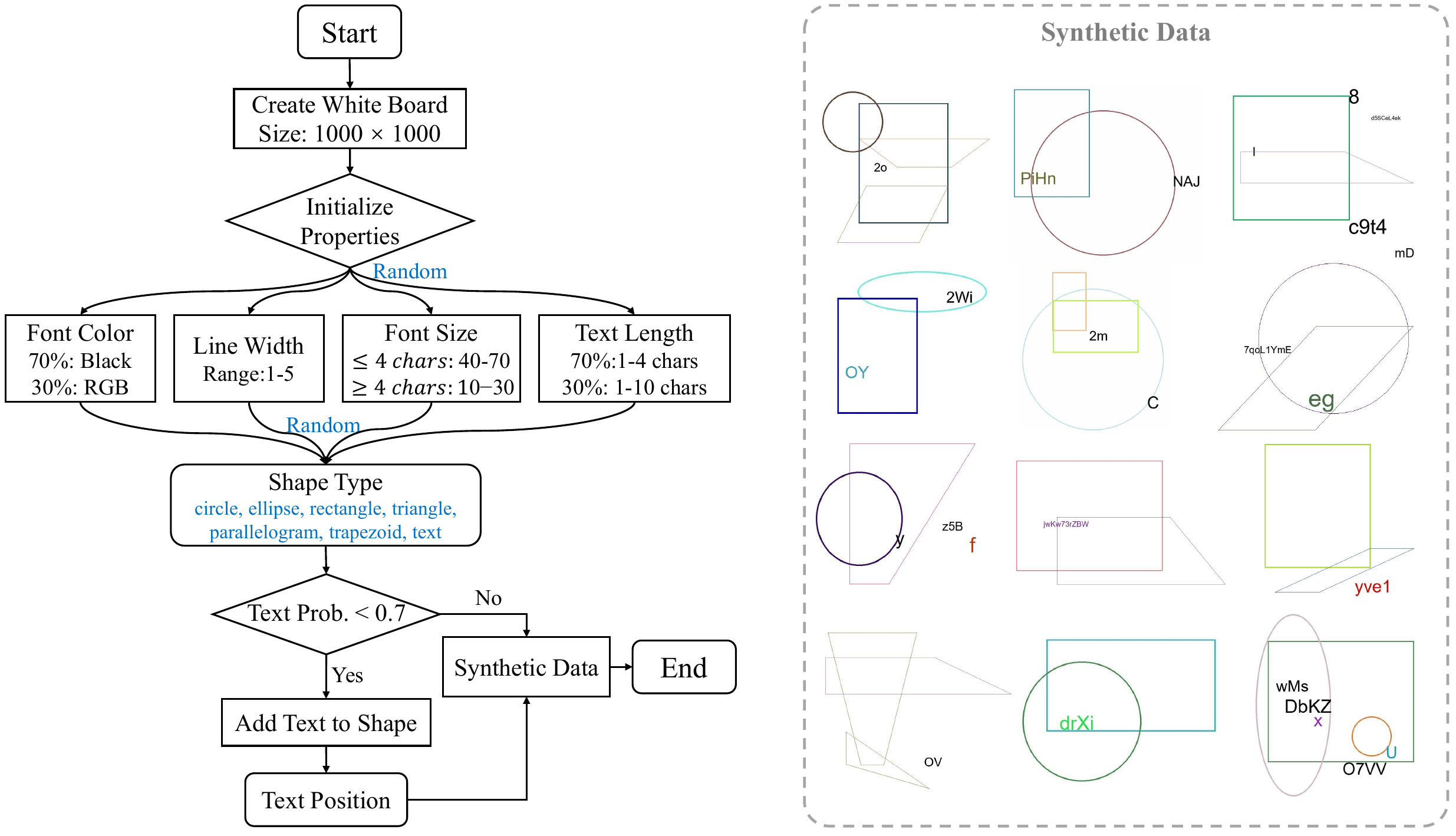}\vspace{-0.2cm}
    \figcaption{The flow diagram depicts the process for generating synthetic math-specific datasets, along with visualizations of the generated data samples.\vspace{-0.3cm}}
    \label{supp:prorgam}
\end{figure}
\subsection{Training Dataset for GeoGLIP }
\label{syndata}
Notably, our synthetic math-specific datasets diffies from the traditional mathematical instruction datasets, and we do not create or use any additional self-generated instruction datasets beyond the publicly available Geo170K~\citep{gao2023g} and MathV360K~\citep{shi2024math} datasets for MLLM training. Instead, our synthetic samples, annotated with box/pixel-level details, are exclusively utilized to train the \ourgip.  Compared to constructing mathematical instruction datasets, our synthetic data generation process is significantly more efficient and resource-friendly. It does not require manual labeling, as all data can be programmatically generated, \eg, through the Matplotlib Python library. In contrast, constructing instruction datasets often relies on GPT-4o to create diverse prompts and necessitates human intervention, making the process labor-intensive and costly.

\noindent\textbf{Shape grounding.} To generate \textit{synthetic datasets} for object grounding tasks, we employ an automated Python-based approach that efficiently creates images containing geometric shapes and text with associated bounding boxes, class labels, and annotations. The geometric categories include shapes like circles, ellipses, rectangles, triangles, parallelograms, trapezoids, and text. A variable number of basic geometric shapes and alphanumeric text elements are generated, with font sizes dynamically adjusted according to text length. These shapes are randomly distributed within a 1000$\times$1000 pixel canvas, while text is positioned either inside or adjacent to the shapes with a 70\% probability. Bounding boxes are then calculated for each shape and text element, ensuring they remain within image bounds. Finally, shapes and text are assigned class labels and coordinates, saved in a COCO-style JSON file for seamless integration with standard GLIP. Fig. \ref{supp:prorgam} shows the detailed flow diagram. Fig. \ref{fig:supp_subfig2} and Fig. Fig. \ref{fig:supp_subfig3}  present the data statistics for synthetic math-specific datasets, including the distribution of geometric shapes and the number of objects per image. Additionally, we incorporated 20,672 images from the \textit{FigureQA} training dataset with bounding box annotations for the shape grounding task.


\noindent\textbf{Junction and boundary detection.} We utilized off-the-shelf models~\citep{huang2018learning,verbin2021field} to extract junctions and boundary as ground truth on both our \textit{synthetic dataset} and public \textit{Geo170K} training images. We then designed junction and boundary heads, parallel to the object detection head, with all tasks sharing the same visual encoder. Through this multi-task learning approach, our \ourgip\ can perceive rich visual information in the mathematical domain.
\subsection{Quantitative analysis}
\label{detevis}
\begin{figure}[t]
    \hspace{-0.8cm}
    \includegraphics[width=1.1\linewidth,height=0.8\linewidth]{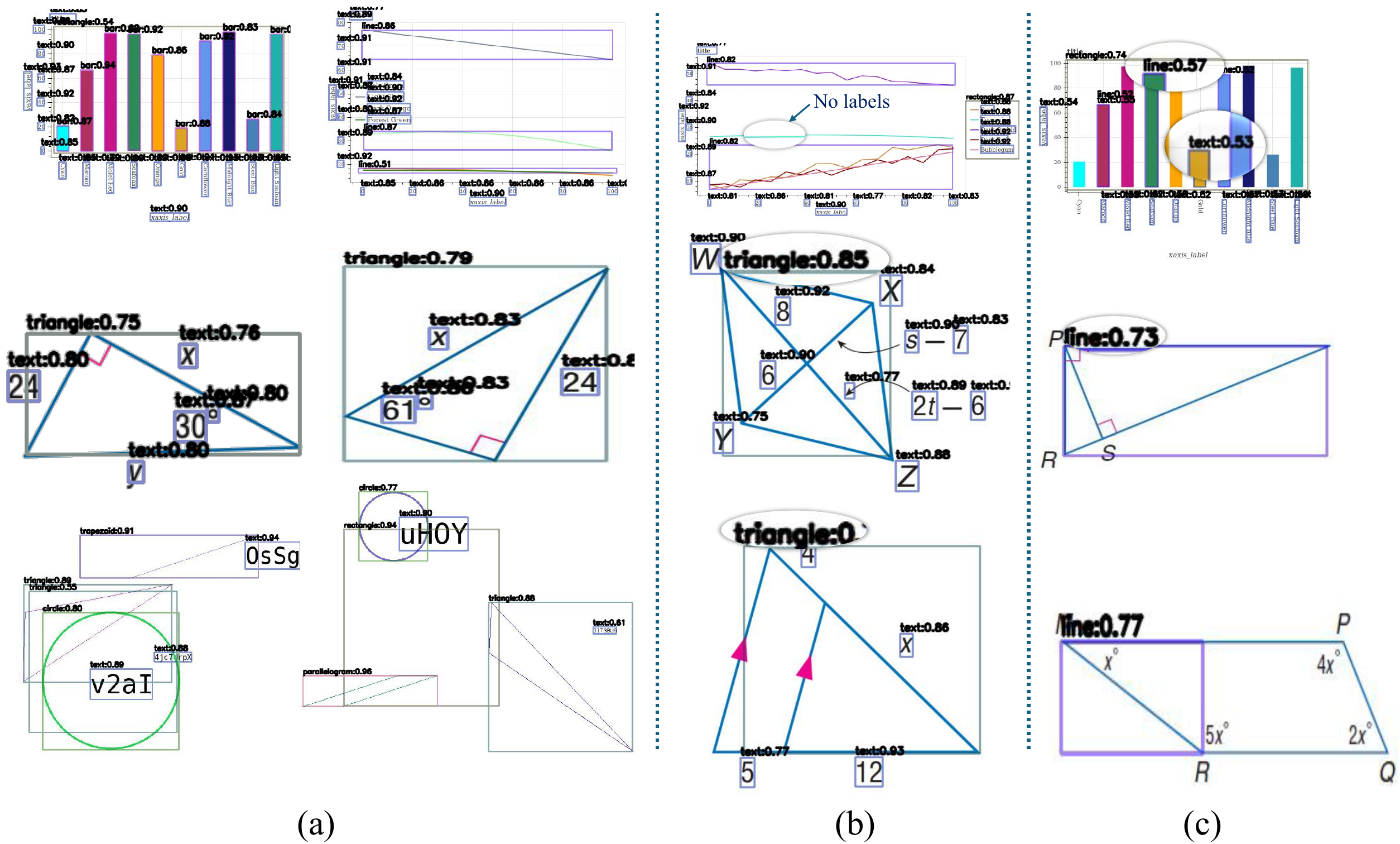}
    \figcaption{The visualization of shape detection on FigrueQA, Geo170K and our synthetic test dataset.  The left panel (a) displays accurate shape detection results generated by \ourgip\, where even small-scale x-ticks are correctly recognized (zoom in 280\% for details). \ourgip\ successfully classifies bars in histograms and rectangular shapes in geometric diagrams. The middle panel (b) represents failure cases, with all errors highlighted using a magnifying glass. For instance, in the first row figure,  the cyan line is misrecognized, and three crowded lines are incorrectly grouped within a single bounding box. The results in the last panel (c) are generated by the original GLIP, trained on natural images. It is evident that most geometric shapes are misclassified as lines or text, and GLIP struggles to recognize small-scale objects, where \ourgip\ excels.}
    \label{fig:devis}
\end{figure}

\begin{figure}[t]
    \centering
    \includegraphics[width=1.\linewidth]{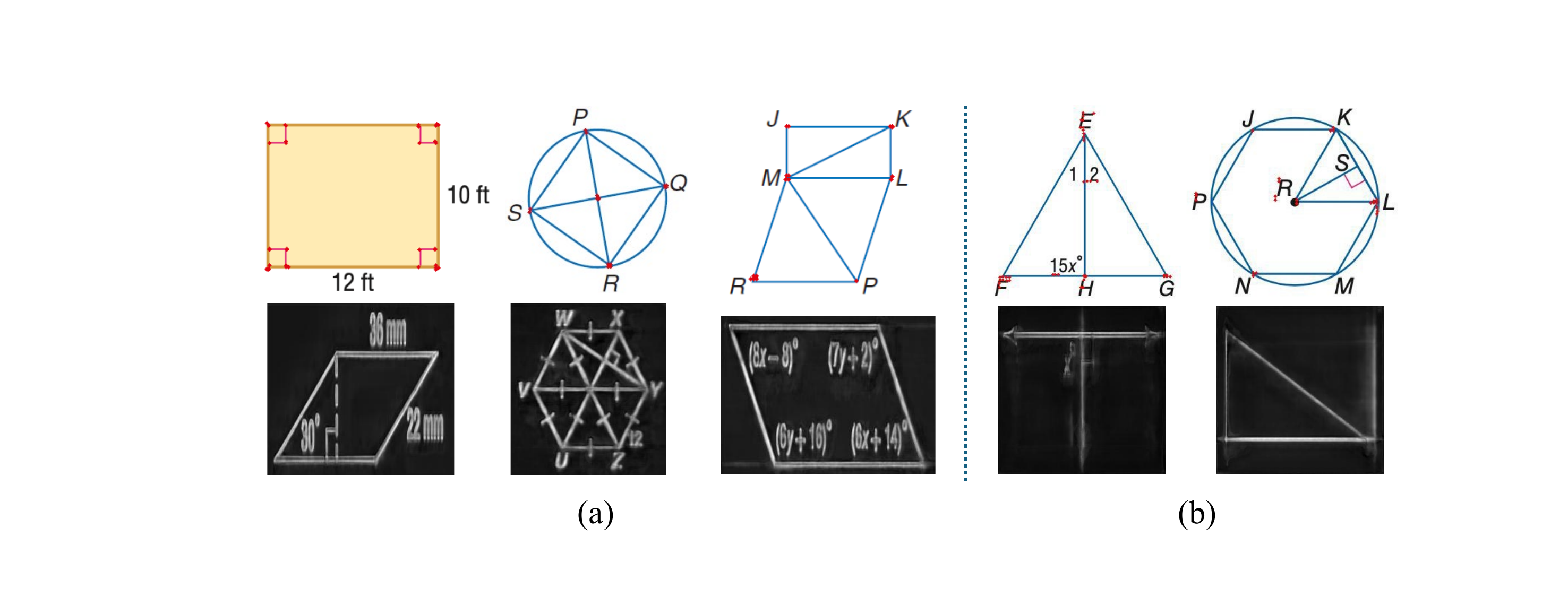}
    \figcaption{The visualization of junction and boundary detection results. The left panel (a) illustrates accurate detections, while the right panel (b) represents failure cases. Junction detection failures frequently exhibit redundant detections.}
    \label{fig:devis2}
\end{figure}

\noindent\textbf{GeoGLIP detection visualizations.} 
Fig.~\ref{fig:devis} illustrates shape detection results on Geo170K, FigureQA and our synthetic test dataset, while Fig.~\ref{fig:devis2} presents the results for boundary and junction detection. Our detector successfully localizes basic geometric shapes and junction points while providing pixel-level boundary results in most cases. However, in complex scenarios such as overcrowded or occluded settings, the detector may struggle. Moreover, in junction detection, some failure cases involve numerous detections but with low accuracy. This issue arises due to noisy ground truth during the training phase, as manually labeling junctions is tedious and time-consuming. To address this, we use an off-the-shelf model~\citep{huang2018learning} to generate ground-truth labels for junction detection. However, since this model was trained on images of man-made environments, it faces an out-of-domain challenge when applied to geometric objects, resulting in labels that are not fully accurate. Improving the accuracy of these labels would significantly enhance junction detection performance.

\noindent\textbf{Providing geometric-relevant information as text inputs.} We have conducted experiments for directly providing geometric-relevant information to the model. Since no existing mathematical instruction datasets include detailed location information for geometric objects (\eg, bounding box coordinates or junction points), we generated this data by inferring Geo170K training images using GeoGLIP to extract the relevant location information. This information was appended to the special token $\langle image \rangle$ in huam value supplementary descriptions for each image, using instructions such as: ``there is a bounding box at $\langle x,y,w,h \rangle$  or there is a junction at $\langle x,y \rangle$  with lines directions $\langle \theta \rangle$''. When tested on the Geo170K test set of the GeoQA benchmark, the top-1 accuracy dropped from 67.0\% to 63.2\%. This result is close to the variant of our constant router 62.8\% (assigning equal weights to all features in Table \ref{rounter}). This performance drop is consistent with our systematic analysis in Fig. \ref{fig:intro_subfig2} and Fig. \ref{fig:intro_subfig3}: Inaccurate instructions would harm the performance, and relevance is key—excessive visual cues interfere with problem-solving.
\subsection{Case studies}
\label{case}
\begin{figure}[t]
    \centering
    \includegraphics[width=1.\linewidth,height=0.7\linewidth]{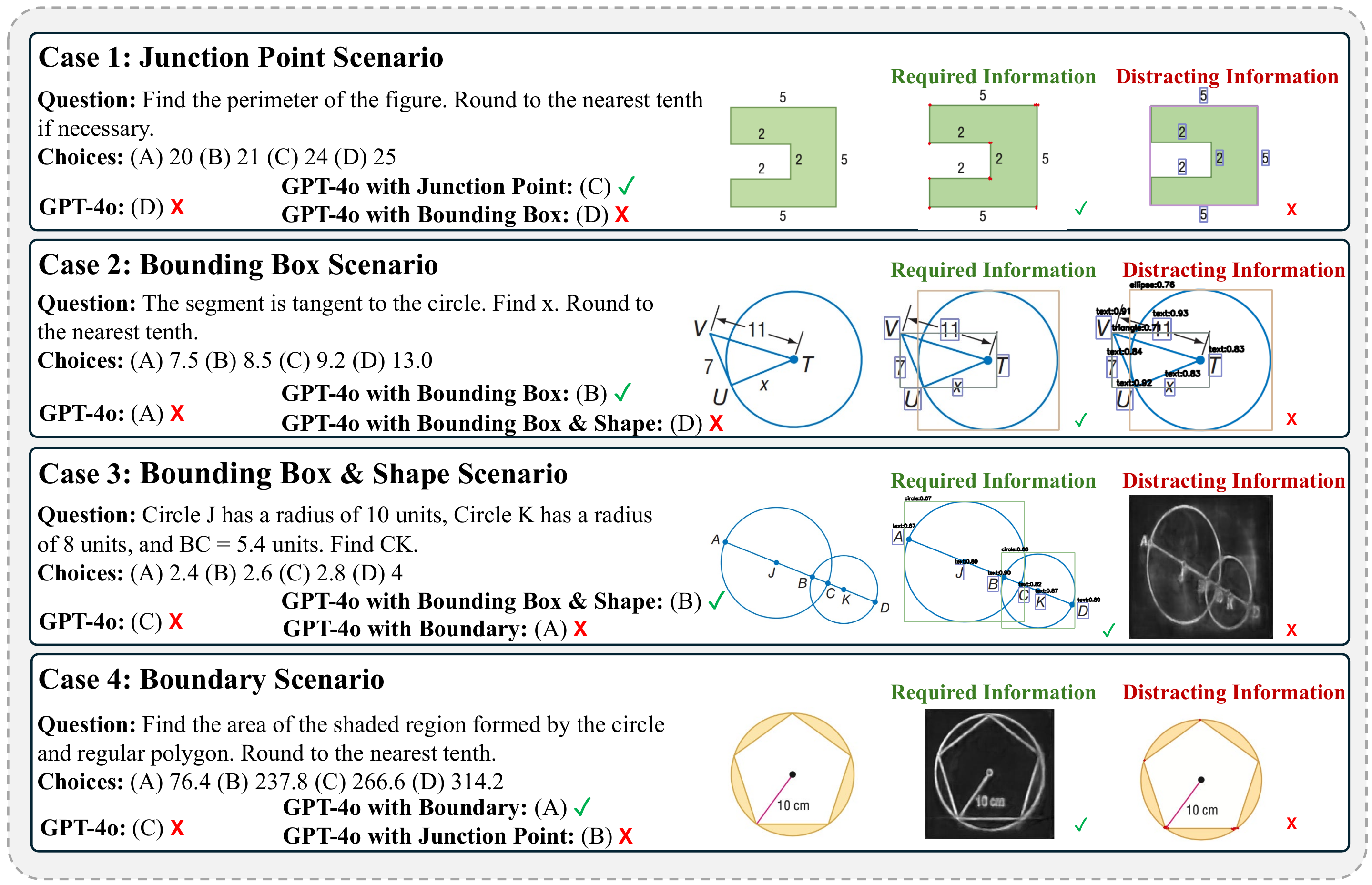}\vspace{-0.2cm}
    \figcaption{{A case study on the Geo170K dataset~\citep{gao2023g} highlights the importance of providing relevant visual information for each math visual question answer. Zoom in for best view}.
    \vspace{-0.1cm}}
    \label{fig:demo}
\end{figure}

\noindent\textbf{Selective visual information helps reasoning.} Fig.~\ref{fig:demo} showcases GPT-4o's responses based on additional visual information from geometric primitives, alongside the question, choices, and diagram $\langle  \text{image} \rangle $  as inputs. We provide hard-coded coordinates for bounding boxes and junctions using instructions such as: ``there is a bounding box at $\langle x,y,w,h \rangle$ (the normalized center point and width/height)" with shape names $\langle \text{geometric shape}\rangle$ (if shape information is provided), or ''candidate junction point $\langle x, y \rangle$. For boundary information, we use ``$\langle  \text{boundary image} \rangle $ is the boundary sketch related to the main diagram" as instructions. The right side visualizes the provided visual cues in the original geometric diagram for clarity, though these images are not input into GPT-4o.  Fig.~\ref{fig:demo} highlights the importance of providing relevant visual prompts for each case; otherwise, redundant information may interfere with the solving process. For example, in case 1, bounding box coordinates per object can be distracting when solving a perimeter question compared to junction locations. In contrast, pixel-level visual information (boundary) aids the model in perceiving complex geometric shapes, such as polygons and circles, and is beneficial for calculating overlap regions, while relying on junctions may lead to biased answers. 
In practice, selecting supporting information for each case is labor-intensive and requires the involvement of math experts. We address this challenge by using the feature router, which automatically learns which fine-grained visual information is important during the training stage. 

Notably, we do not claim that the feature router can explicitly select specific types of visual information, such as bounding boxes, junctions, or shapes. This is because the inputs to the feature pyramid of the \ourgip\ visual encoder do not clearly represent each type of information in a distinct manner. Since \ourgip\ is trained on multiple tasks using a shared visual encoder, it becomes challenging to determine which specific feature maps correspond to which an individual learning task. What our findings emphasize is the importance of selecting optimal visual cues, demonstrating that while accuracy is crucial, more information does not always lead to better performance—relevance is key. We anticipate that more advanced selection techniques could further enhance mathematical problem-solving in visual contexts. Refer to Sec.~\ref{furwork} for our future research directions.

\begin{figure}[t]
    \centering
    \begin{subfigure}[b]{0.48\linewidth}
        \centering
        \includegraphics[width=1\linewidth]{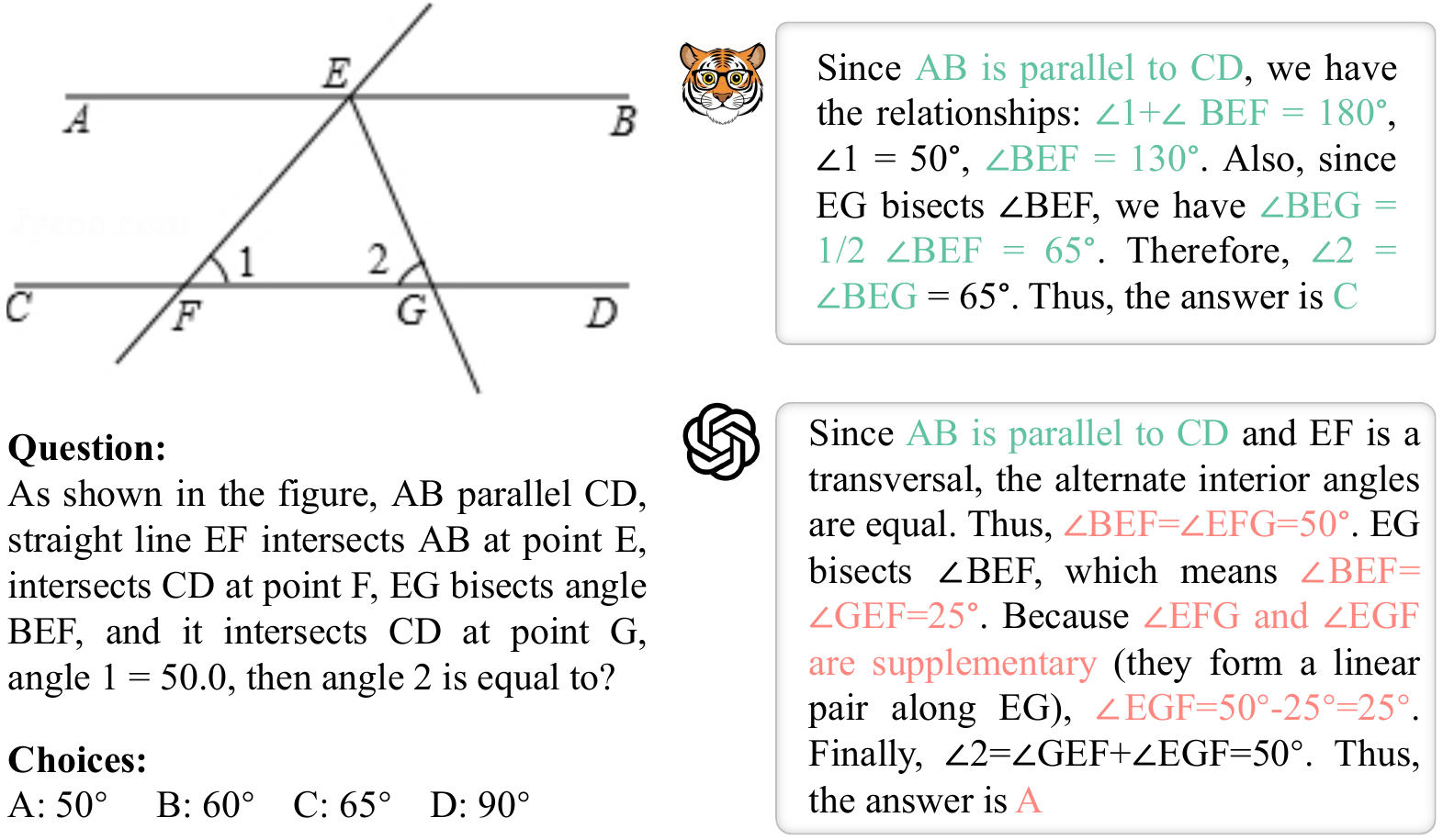}
    \caption{}
    \label{fig:cap1}
    \end{subfigure}
    \hfill
    \begin{subfigure}[b]{0.48\linewidth}
        \centering
        \includegraphics[width=\linewidth]{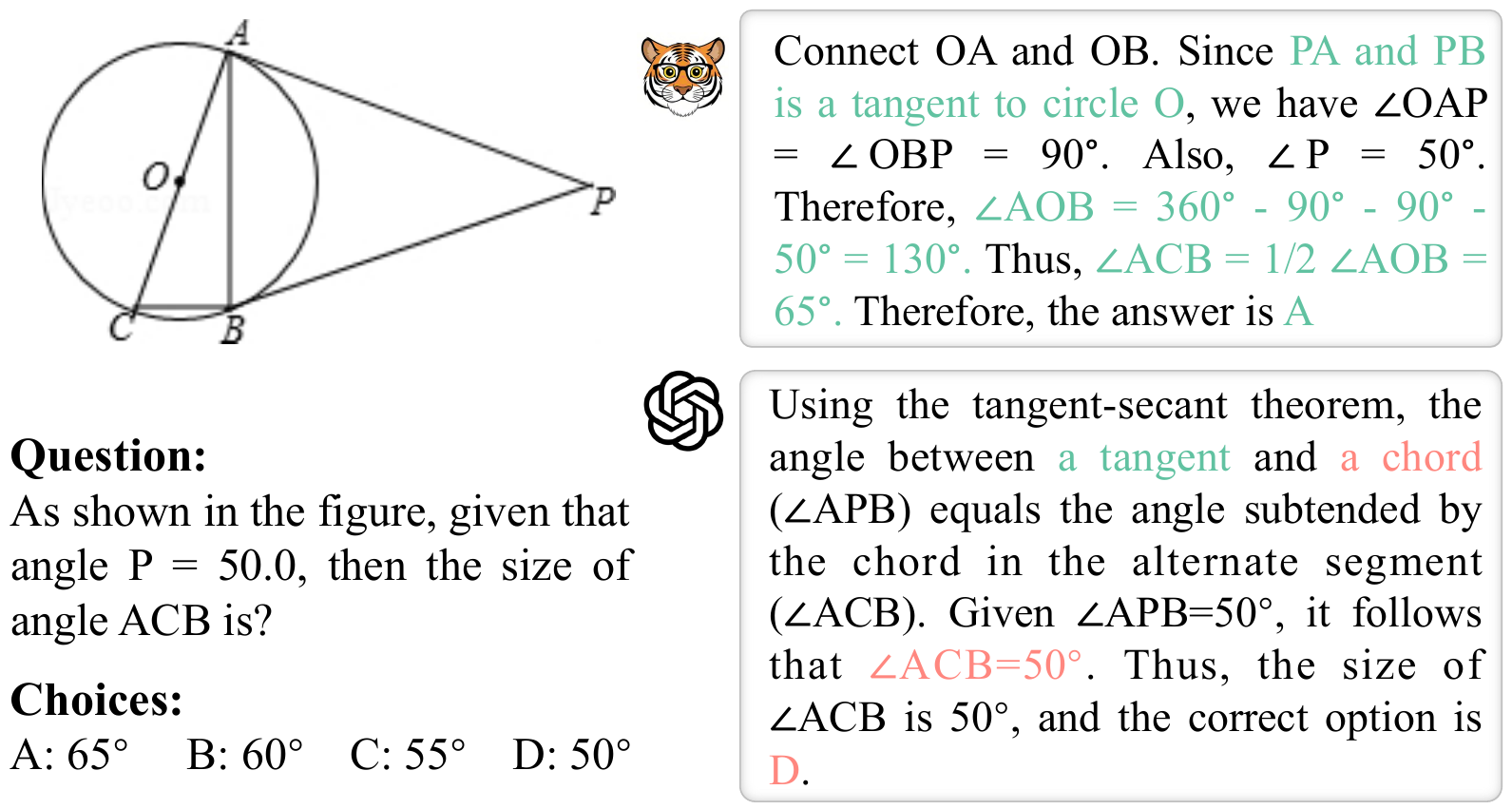}
    \caption{}
    \label{fig:cap2}
    \end{subfigure}
    
    \caption{Response comparison of our \ourMethod-Deepseek-7B and GPT-4o. Refer to the main text for detailed analysis. Zoom in for best view.}
    \label{fig:cap}
\end{figure}

\noindent\textbf{Response comparison.} 
Fig. \ref{fig:cap} presents case studies comparing our \ourMethod-Deepseek-7B with GPT-4o on the MathVerse testmini set. These examples highlight the strengths of \ourMethod-Deepseek-7B in providing precise geometric visual information, enabling clear and logically grounded mathematical reasoning in its responses.
For instance, our model demonstrates sensitivity to the positions of individual points/junctions, effectively capturing the relationships between different lines. As shown in Fig. \ref{fig:cap1}, it successfully identifies angle 1 and its relationship with angle BEF, enabling correct reasoning and answers. In contrast, GPT-4o fails to recognize these relationships, leading to flawed reasoning and incorrect answers.

\begin{figure}[t]
    \centering
    \includegraphics[width=1\linewidth]{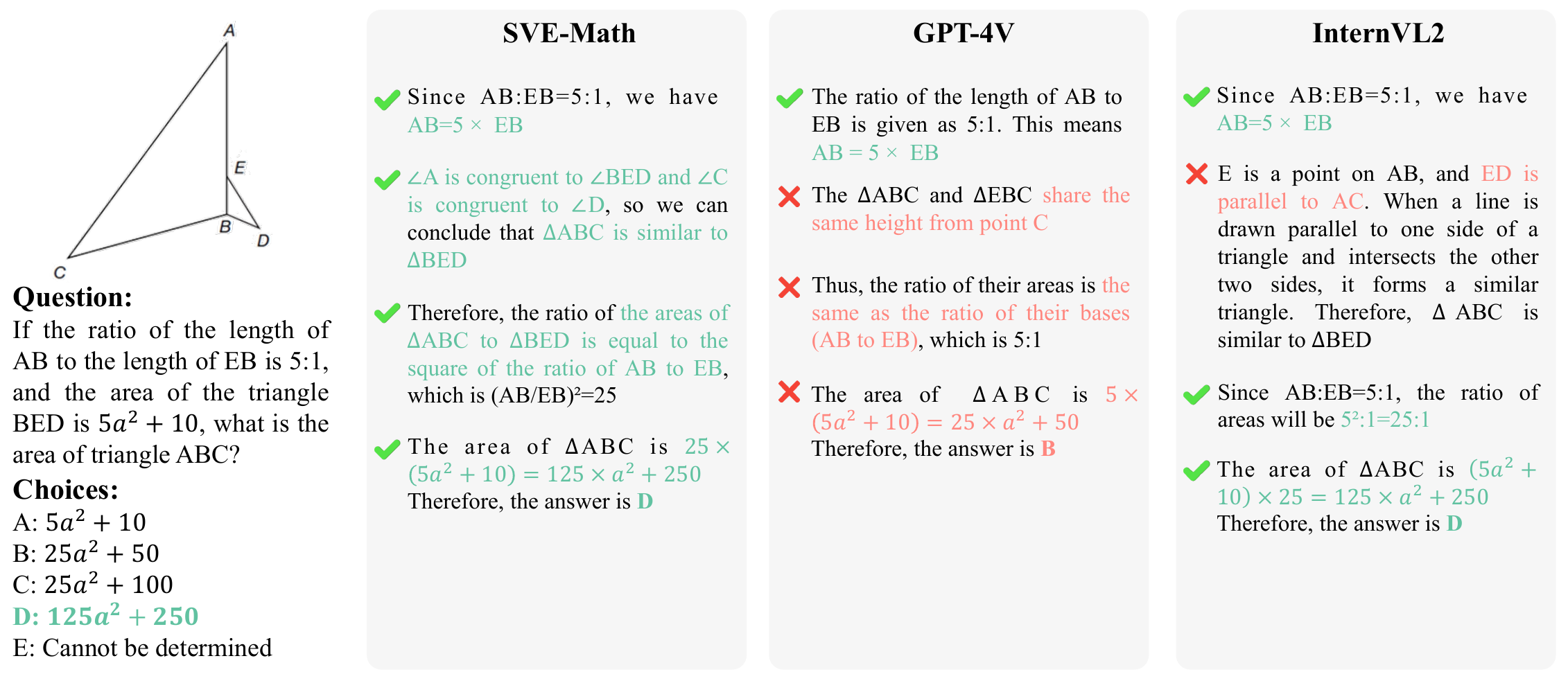}\vspace{-0.2cm}
    \figcaption{{Chain-of-Thought (CoT) response comparison of our \ourMethod-Deepseek-7B, GPT-4V and InternVL2. Refer to the main text for detailed analysis. Zoom in for best view.}}
    \label{fig:cot1}
\end{figure}

\begin{figure}[t]
    \centering
    \includegraphics[width=1\linewidth]{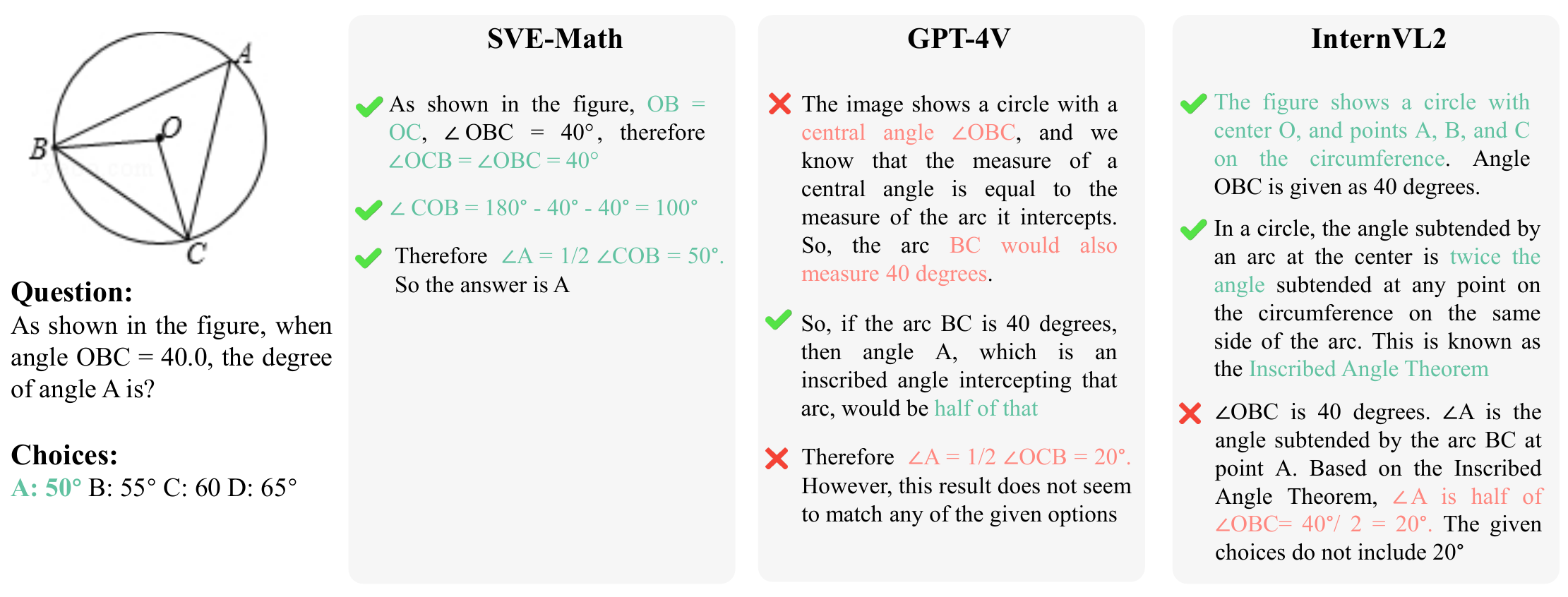}\vspace{-0.2cm}
    \figcaption{{ Chain-of-Thought (CoT) response comparison of our \ourMethod-Deepseek-7B, GPT-4V and InternVL2. Refer to the main text for detailed analysis. Zoom in for best view.}}
    \label{fig:cot2}
\end{figure}

Fig. \ref{fig:cot1} and Fig. \ref{fig:cot2} present a Chain-of-Thought (CoT) comparison among \ourMethod-Deepseek-7B, GPT-4V, and InternVL2. The results clearly demonstrate that providing geometry-aware visual cues significantly aids LLMs in understanding the relationships between geometric elements, thereby enhancing the entire reasoning process. In contrast, the other two MLLMs fail to achieve this level of understanding, leading to incorrect reasoning and outcomes. This demonstrates that without accurately recognizing visual elements, even strong LLMs struggle with reasoning tasks. As shown in GPT-4V's output, its initial misidentification of mathematical elements results in an incorrect Chain-of-Thought (CoT) response.
 
\subsection{Mathematical Visual Training and Efficiency} 
\label{cost}
\subsubsection{Training Details} 
\label{train}
Our work follows a structured three-stage training pipeline, including multi-task visual perception training for \ourgip, visual-language alignment, and mathematical instruction tuning for MLLMs.

\noindent \textbf{Stage 1:} To enable the visual encoder in \ourgip\ to ground geometric entities in mathematical diagrams, we utilize synthetic and FigureQA training images annotated with bounding boxes for the \textit{grounded pre-training}. Specifically, we fine-tune a pre-trained GLIP-T model (with Swin-Tiny as the backbone), adhering to the GLIP detection loss defined as:
\begin{equation}
\mathcal{L}_{det} = \mathcal{L}_{rpn} + \mathcal{L}_{cls} + \mathcal{L}_{reg}
\label{equ:loss_det}
\end{equation}

where $\mathcal{L}_{rpn}$ refines the region proposals generated by the RPN, $\mathcal{L}_{cls}$ applies binary sigmoid loss to alignment scores, and $\mathcal{L}_{reg}$ uses smooth $\ell_1$ loss for bounding box regression. 

Following the process in~\citep{huang2018learning}, for the \textit{junction detection} task, the input image is divided into mesh grids, with each grid cell responsible for detecting a junction if its center falls within the cell. Each $ij$-th cell predicts a confidence score $c_{ij}$, indicating the likelihood of a junction in that cell. Since a junction represents the intersection of lines, the number of predictions per cell varies depending on the number of lines intersecting. To capture orientations, each cell is further divided into $K$ equal bins (default $K=15$), with each bin spanning 24 degrees to cover the full 360-degree range.
Each junction is represented as $JP_{ij} = (x_{ij}, c_{ij}, \{\theta_{ijk}, c^{\theta}_{ijk}\}_{k=1}^{K})$, where $x_{ij}$ denotes the junction center coordinates, $c_{ij} \in [0,1]$ is the confidence score for the presence of a junction, $\theta_{ijk}$ is the angle of the $k$-th bin, and $c^{\theta}_{ijk}$ is the confidence score for that bin. 

The loss function for \text{junction detection} consists of four terms. Given a set of ground truth junctions $JP = {jp_1, \dots, jp_N}$ in an image, the loss function is formulated as:
\begin{equation}
 \mathcal{L}_{junc} = \lambda_{loc} \cdot (\mathcal{L}^{c}_{{loc}} + \mathcal{L}^b_{{loc}}) + \lambda_{conf} \cdot (\mathcal{L}^b_{{conf}} +\mathcal{L}^b_{conf}).
\label{equ:loss_jun}
\end{equation}

The default values for the weights in Eq.~\ref{equ:loss_jun} are $\lambda_{loc}=0.1$ and $\lambda_{conf}=1$, where the superscripts $c$ and $b$ refer to cell and bin, respectively. Specifically, we apply the binary cross-entropy loss for both $\mathcal{L}^{c}_{conf}$ and $\mathcal{L}^{b}_{conf}$, and use $\ell_2$ loss to measure the relative position of the predictions against the ground truth for $\mathcal{L}^{c}_{loc}$ and $\mathcal{L}^{b}_{loc}$. Refer to~\citep{huang2018learning} for more details. In the \textit{boundary detection} task, $\mathcal{L}_{bodr}$ minimizes the $\ell_2$ loss between the estimated heatmap values and the ground truth values.

Our final loss function for multi-task visual perception training is defined as:
\begin{equation}
\mathcal{L}_{\text{vis}} = \mathcal{L}_{{det}} + \mathcal{L}_{{junc}} + 5 \cdot \mathcal{L}_{{bodr}},
\label{equ:loss_vis}
\end{equation}
where the weight for $\mathcal{L}_{\text{bodr}}$ is set to 5, while the weights for $\mathcal{L}_{\text{det}}$ and $\mathcal{L}_{\text{junc}}$ are kept at 1. 

\noindent \textbf{Stage 2 \& 3:}  During both phases, we freeze the \ourgip\ encoder. In Stage 2, we train only the projection layers to align diagram-language pairs. In Stage 3, we unfreeze both the projection layer and the LLM to perform comprehensive instruction-following tuning, culminating in \ourMethod-7B. For these two stages, we employ the conventional LLaVA loss, formulated as:
\begin{equation}
\mathcal{L}_{llm} = - \sum_{t=1}^{L} \log p \left[ S_{tar}^t | f_{\phi}(s_{tar}^{(<t)}, S_{in}, I) \right],
\label{equ:loss_llm}
\end{equation}
where $f_{\phi}$ denotes the model parameterized by $\phi$, $I$ corresponds to the figure, $S_{tar}$ and $S_{in}$ represent the target and input sentences, respectively; $S_{tar}^t$ refers to the $t$-th token of the target output, and $L$ denotes the sequence length.

\subsubsection{Efficiency}
\ourMethod-7B introduces minimal computational overhead, as detailed in the below comparison Table \ref{tab:parameter_comparison}. The \ourgip\ and Connector contribute an additional parameter size of 32.65MB and 8.73MB, and the Projectors accounting for 16.13MB. The inference time per sample increases slightly, from 19.80s to 20.04s (+0.24s). Training is conducted on 8 A100 GPUs with a batch size of 128 using the MathV360K dataset, which includes 40K images and 360K question-answer pairs. The total training time shows only a marginal increase, from 10.35h to 10.54h (+0.19h), demonstrating scalability for larger models and datasets.
\begin{table}[h!]
\centering
\caption{Comparison of computational overhead and parameter size for G-LLaVA and SVE-Math.}
\begin{adjustbox}{width=1\linewidth}
\begin{tabular}{|c|c|c|c|c|c|}
\hline
\textbf{\#Parameter size} & \textbf{GeoGLIP} & \textbf{Connector} & \textbf{Projectors} & \textbf{Time (inference/sample)} & \textbf{Time (training/MathV360K)} \\ \hline
G-LLaVA & - & - & 16.52MB & 19.80s & 10.35h \\ \hline
\textbf{\ourMethod} & 32.65MB & 8.73MB & 31.20MB & 20.04s & 10.54h \\ \hline
\end{tabular}
\end{adjustbox}
\vspace{-0.5cm}
\label{tab:parameter_comparison}
\end{table}

\subsection{Limitations and Further research}
\label{furwork}
Our research aims to offer a new perspective on solving mathematical visual reasoning problems by first training a vision-centric model to provide visual prompts for LLMs, rather than focusing on creating large visual instruction fine-tuning datasets for MLLMs. Despite the effectiveness of our approach, there are several limitations to consider. First, the reliance on synthetic data for visual tasks may not fully capture the complexity of real-world geometric problems, potentially limiting generalization to more diverse datasets. Additionally, while the feature router provides automatic selection of relevant visual cues, it may not always perfectly align with human intuition or domain-specific knowledge, particularly in cases requiring more intricate reasoning.

For future research, one promising direction is to extend our method by incorporating real-world mathematical datasets to improve generalization and robustness. However, this will require some human labeling processes, as existing mathematical datasets lack detailed box or pixel-level annotations. Incorporating such annotations would provide a more accurate and fine-grained understanding of visual elements in mathematical problems, allowing models to better generalize to real-world scenarios. Developing efficient semi-automated labeling techniques or combining expert annotations with synthetic data could also help reduce the manual effort required. With improved detection performance, we may explore more advanced methods for designing soft prompts, such as object-level prompts.  
Further refinement of the feature router, such as combining it with interpretable methods to better understand its decision-making process, could also enhance the model's performance. By making the feature router more transparent, we could gain insights into how it selects and prioritizes visual cues, allowing for fine-tuning that aligns better with human intuition and task-specific requirements. This, in turn, would allow for more informed adjustments, leading to better generalization and accuracy in complex mathematical problem-solving scenarios.

\begin{figure}[t]
    \centering
    \includegraphics[width=0.8\linewidth]{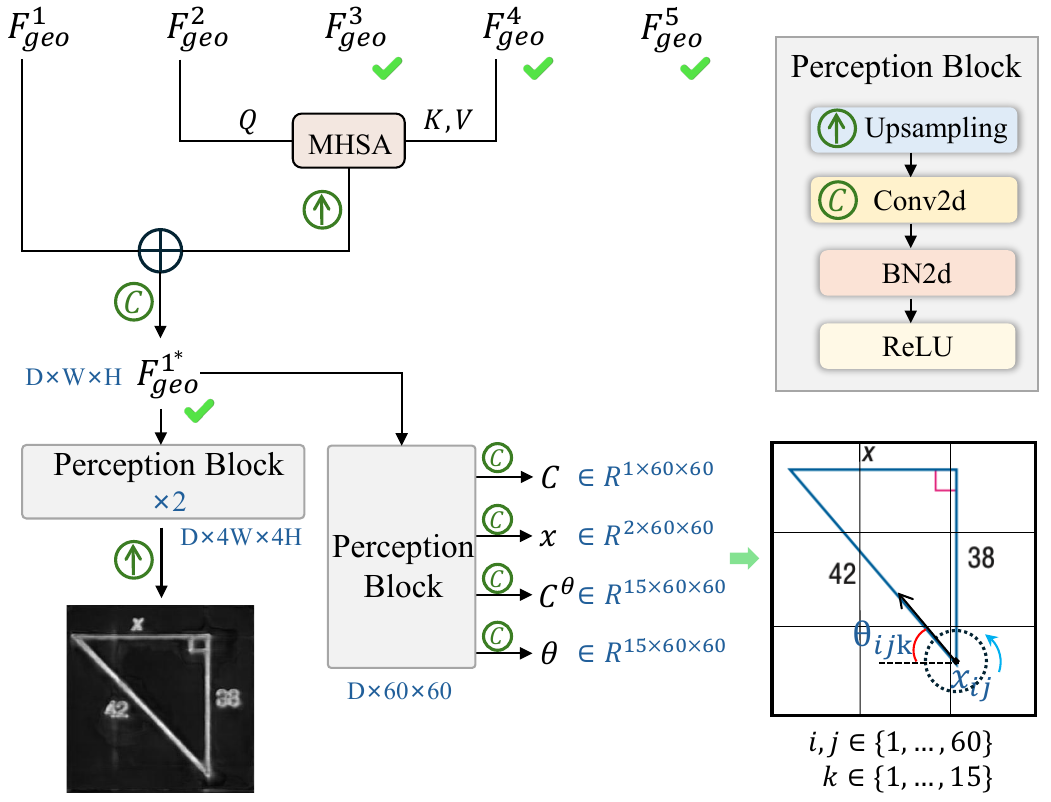}\vspace{-0.2cm}
    \figcaption{Designs for the junction and boundary detectors: We first use an attention mechanism (MHSA) to fuse two-scale features, followed by upsampling and addition with the highest resolution features, resulting in $F_{\text{geo}}^{1^*}$. Separate perception blocks are then applied for junction and boundary detection. For junction detection, the detector provides cell confidence ($C$), cell location ($X$), bin confidence ($C^\theta$), and bin orientation ($\theta$). Green check-marked features indicate candidate features for soft prompts, with $D,W,H$ representing channel dim., and spatial resolution (width\&height). \vspace{-0.1cm}}
    \label{fig:detail}
\end{figure}

\end{document}